\documentclass[10pt,journal,cspaper,compsoc]{IEEEtran}
\ifCLASSOPTIONcompsoc
\else
\fi

\ifCLASSINFOpdf
\else
\fi

\usepackage{graphicx} 
\usepackage{subfigure}

\usepackage[numbers]{natbib}

\usepackage{algorithm}
\usepackage{algorithmic}

\usepackage{amssymb}
\usepackage{amsmath}
\usepackage{enumerate}
\usepackage{epsf}
\usepackage{algorithm}
\usepackage{algorithmic}
\usepackage{wrapfig}
\usepackage{sidecap}
\usepackage{eqnarray}
\usepackage{multirow}
\usepackage{color}

\usepackage{graphics} 

\newcommand{\xv}{\mathbf{x}}

\newcommand{\Zv}{\mathbf{Z}}

\newcommand{\ev}{\mathbf{e}}

\newcommand{\sign}{\mathrm{sign}}

\newcommand{\nuv}{\boldsymbol \nu}
\newcommand{\psiv}{\boldsymbol \psi}
\newcommand{\xiv}{\boldsymbol \xi}

\newcommand{\piv}{\boldsymbol \pi}
\newcommand{\omegav}{\boldsymbol \omega}
\newcommand{\gammav}{\boldsymbol \gamma}

\newcommand{\ep}{\mathbb{E}}

\newcommand{\KL}{\mathrm{KL}}
\newcommand{\B}{\mathrm{Beta}}

\newcommand{\Ber}{\mathrm{Bernoulli}}

\newcommand{\data}{\mathcal{D}}

\newcommand{\model}{\mathcal{M}}
\newcommand{\ms}{\mathbb{M}}
\def\indicator{{\mathbb I}}

\newcommand{\junz}[1]{{\color{blue}{\bf\sf [JZ: #1]}}}
\newcommand{\jun}[1]{{\color{red}{\bf\sf [#1]}}}

\newcommand{\jiaming}[1]{{\color{blue}{\bf\sf [JM: #1]}}}

\begin{document}
%
\title{Max-Margin Nonparametric Latent Feature Models for Link Prediction}
%
%
%
%

\author{Jun Zhu,~\IEEEmembership{Member,~IEEE}, ~ Jiaming Song, ~ Bei Chen
\IEEEcompsocitemizethanks{\IEEEcompsocthanksitem J. Zhu, J. Song and B. Chen are with the Department
of Computer Science and Technology, State Key Lab of Intelligent Technology and Systems, Tsinghua National Lab for Information Science and Technology, Tsinghua University, Beijing, 100084 China.\protect\\
E-mail: dcszj@mail.tsinghua.edu.cn;~ jiaming.tsong@gmail.com;~ beichen1019@126.com}
\thanks{}}

\markboth{Journal of \LaTeX\ Class Files,~Vol.~6, No.~1, January~2007}%
{Shell \MakeLowercase{\textit{et al.}}: Bare Demo of IEEEtran.cls for Computer Society Journals}
%

\IEEEcompsoctitleabstractindextext{%
\begin{abstract}

Link prediction is a fundamental task in statistical network analysis. Recent advances have been made on learning flexible nonparametric Bayesian latent feature models for link prediction. In this paper, we present a max-margin learning method for such nonparametric latent feature relational models. Our approach attempts to unite the ideas of max-margin learning and Bayesian nonparametrics to discover discriminative latent features for link prediction. It inherits the advances of nonparametric Bayesian methods to infer the unknown latent social dimension, while for discriminative link prediction, it adopts the max-margin learning principle by minimizing a hinge-loss using the linear expectation operator, without dealing with a highly nonlinear link likelihood function. For posterior inference, we develop an efficient stochastic variational inference algorithm under a truncated mean-field assumption. Our methods can scale up to large-scale real networks with millions of entities and tens of millions of positive links. We also provide a full Bayesian formulation, which can avoid tuning regularization hyper-parameters. Experimental results on a diverse range of real datasets demonstrate the benefits inherited from max-margin learning and Bayesian nonparametric inference.

\end{abstract}

}

\maketitle

\IEEEdisplaynotcompsoctitleabstractindextext

%
\IEEEpeerreviewmaketitle

\section{Introduction}
%
%

\IEEEPARstart{A}{s} the availability and scope of social networks and relational datasets increase in both scientific and engineering domains, a considerable amount of attention has been devoted to the statistical analysis of such data,
which is typically represented as a graph with the vertices denoting entities and edges denoting links between entities.
Links can be either undirected (e.g., coauthorship on papers) or directed (e.g., citations).
Link prediction is a fundamental problem in analyzing these relational data, and its goal is to
predict unseen links between entities given the observed links. Often there is extra information about links
and entities such as attributes and timestamps~\cite{liben_nowell,backstrom,Miller:nips09} that can be used to help with prediction.

Link prediction has been examined in both unsupervised and supervised learning settings, while supervised methods often have better results~\cite{Hasan:2006,Lichtenwalter:2010}. Recently, various approaches based on probabilistic latent variable models have been developed.
One class of such models utilize a latent feature matrix and a link function (e.g., the commonly used sigmoid function)~\cite{hoff, Miller:nips09} to
define the link formation probability distribution. These {\it latent feature models} were shown to
generalize latent class~\cite{Nowicki:2001,Airoldi:nips08} and latent distance~\cite{Hoff:02} models and are thus able
to represent both homophily and stochastic equivalence, which are important properties commonly observed in real-world
social network and relational data. The parameters for these probabilistic latent variable models are typically estimated with an EM algorithm to do maximum-likelihood estimation (MLE)
or their posterior distributions are inferred with Monte Carlo methods under a Bayesian formulation. Such techniques have
demonstrated competitive results on various datasets. However, to determine the unknown dimensionality of the latent feature space (or latent social space),
most of the existing approaches rely on an external model selection procedure, e.g., cross-validation, which could be expensive by comparing many different settings.

Nonparametric Bayesian methods~\cite{Orbanz:2010} provide alternative solutions, which bypass model selection by inferring the model complexity from data in a single learning procedure. The nonparametric property is often achieved by using a flexible prior (a stochastic process) on an unbounded measure space. Popular examples include Dirichlet process (DP)~\cite{Ferguson:73} on a probability measure space, Gaussian process (GP)~\cite{Rasmussen:2006} on a continuous function space, and Indian buffet process (IBP)~\cite{Griffiths:nips06} on a space of unbounded binary matrices that can have an infinite number of columns. For link prediction, DP and its hierarchical extension (i.e., hierarchical Dirichlet process, or HDP)~\cite{Teh:2006} have been used to develop nonparametric latent class models~\cite{Kemp:aaai06,Kim:2013}. For latent feature models, the work~\cite{Miller:nips09} presents a nonparametric Bayesian method to automatically infer the unknown social dimension.

This paper presents an alternative way to develop nonparametric latent feature relational models.
Instead of defining a normalized link likelihood model, we propose to directly minimize some objective function (e.g., hinge-loss)
that measures the quality of link prediction, under the principle of maximum entropy discrimination (MED)~\cite{Jaakkola:MED99,Jebara:thesis01},
an elegant framework that integrates max-margin learning and Bayesian generative modeling.
The present work extends MED in several novel ways to solve the challenging link prediction problem.
First, like~\cite{Miller:nips09}, we use nonparametric Bayesian techniques to automatically resolve the unknown
dimension of a latent social space, and thus our work represents an attempt towards uniting Bayesian nonparametrics and max-margin learning,
which have been largely treated as two isolated topics, except a few recent successful examples~\cite{Zhu:icml11,Zhu:regBayes14,Xu:icml13}.
Second, we present a full Bayesian method to avoid tuning regularization constants.
Finally, by minimizing a hinge-loss, our model avoids dealing with a highly nonlinear link likelihood (e.g., sigmoid) and can be
efficiently solved using variational methods, where the sub-problems of max-margin learning
are solved with existing high-performance solvers.
We further develop a stochastic algorithm that scales up to massive networks.
Experimental results on a diverse range of real datasets demonstrate that 1) max-margin learning can significantly improve the link prediction performance of Bayesian latent feature relational models;
2) using full Bayesian methods, we can avoid tuning regularization constants without sacrificing the performance,
and dramatically decrease running time;
and 3) using stochastic methods, we can achieve high AUC scores on the US Patents network, which consists of
millions of entities and tens of millions of positive links.


The paper is structured as follows. Section 2 reviews related work.
Section 3 presents the max-margin latent feature relational model, with a stochastic algorithm and a full Bayesian formulation.
Section 4 presents empirical results. Finally, Section 5 concludes.

\vspace{-.1cm}
\section{Related Work}
\vspace{-.1cm}


We briefly review the work on link prediction, latent variable relational models and MED.

\vspace{-.2cm}
\subsection{Link Prediction}
\vspace{-.1cm}

Many scientific and engineering data are represented as networks, such as social networks and biological gene networks. Developing statistical models to analyze such data has attracted a considerable amount of attention, where link prediction is a fundamental task~\cite{liben_nowell}. For static networks, link prediction is defined to predict unobserved links by using the knowledge learned from observed ones, while for dynamic networks, it is defined as learning from the structures up to time $t$ in order to predict the network structure at time $t+1$. The early work on link prediction has been focused on designing good proximity (or similarity) measures between nodes, using features related to the network topology. The measure scores are used to produce a rank list of candidate link pairs. Popular measures include common neighbors, Jaccard's coefficient~\cite{Salton:1983}, Adamic/Adar~\cite{Adamic:2003}, and etc. Such methods are unsupervised in the sense that they do not learn models from training links. Supervised learning methods have also been popular for link prediction~\cite{Hasan:2006,Lichtenwalter:2010,Shi:2009}, which learn predictive models on labeled training data with a set of manually designed features that capture the statistics of the network.

\vspace{-.2cm}
\subsection{Latent Variable Relational Models}
\vspace{-.1cm}

Latent variable models (LVMs) have been popular in network analysis as: 1) they can discover latent structures (e.g., communities) of network data; and 2) they can make accurate predictions of the link structures using automatically learned features. Existing LVMs for network analysis can be grouped into two categories---{\it latent class models} and {\it latent feature models}.

Latent class models assume that there are a number of clusters (or classes) and each entity belongs to a single cluster. Then, the probability of a link between two entities depends only on their cluster assignments. Representative work includes stochastic block models \cite{Nowicki:2001} and their nonparametric extensions, such as the infinite relational model (IRM)~\cite{Kemp:aaai06} and the infinite hidden relational model~\cite{Xu:2006}, which allow a potentially infinite number of clusters. Given a dataset, the nonparametric methods automatically infer the number of latent classes. The mixed membership stochastic block model (MMSB)~\cite{Airoldi:nips08} increases the expressiveness of latent class models by allowing each entity to associate with multiple communities. But the number of latent communities is required to be externally specified. The nonparametric extension of MMSB is a hierarchical Dirichlet process relational (HDPR) model~\cite{Kim:2013}, which allows mixed membership in an unbounded number of latent communities.

For latent feature models, each entity is assumed to be associated with a feature vector, and the probability of a link is determined by the interactions among the latent features. 
The latent feature models are more flexible than latent class models, which may need an exponential number of classes in order to be equal on model expressiveness. Representative work in this category includes the latent distance model~\cite{Hoff:02}, the latent eigenmodel~\cite{hoff}, and the nonparametric latent feature relational model (LFRM)~\cite{Miller:nips09}. As these methods are closely related to ours, we will provide a detailed discussion of them in next section.

The expressiveness of latent features and the single-belonging property of latent classes are not exclusive. In fact, they can be combined to develop more advanced models. For example,~\cite{Palla:2012} presents an infinite latent attribute model, which is a latent feature model but each feature is itself partitioned into disjoint groups (i.e., subclusters). In this paper, we focus on latent feature models, but our methods can be extended to have a hierarchy of latent variables as in~\cite{Palla:2012}. 

\vspace{-.2cm}
\subsection{Maximum Entropy Discrimination}
\vspace{-.1cm}

We consider binary classification, where the response
variable $Y$ takes values from $\{+1, -1\}$. Let $X$ be an input feature vector and $F(X; \eta)$ be a discriminant function parameterized by $\eta$.
Let $\data = \{(X_n, Y_n)\}_{n=1}^N$ be a training set and define $h_\ell(x) = \max(0, \ell-x)$, where $\ell$ is a positive cost parameter.
Unlike standard SVMs, which estimate a single $\eta$,
maximum entropy discrimination (MED)~\cite{Jaakkola:MED99} learns a distribution $p(\eta)$ by solving an entropic regularized risk minimization problem with prior $p_0(\eta)$\vspace{-.2cm}
\begin{eqnarray}\label{problem:med}
 \min_{p(\eta)} ~ \KL(p(\eta) \Vert p_0(\eta)) + C \cdot \mathcal{R}(p(\eta)),\vspace{-.3cm}
\end{eqnarray}
where $C$ is a positive constant; $\KL(p\Vert q)$ is the KL divergence;
$\mathcal{R}(p(\eta)) = \sum_n h_1( Y_n \ep_{p(\eta)} [F(X_n; \eta)] )$ is the hinge-loss that
captures the large-margin principle underlying the MED prediction rule\vspace{-.2cm}
\begin{eqnarray}\label{problem:med_rule}
 \hat{Y} = \sign \left( \ep_{p(\eta)} [ F(X; \eta) ] \right).\vspace{-.3cm}
\end{eqnarray}
MED subsumes SVM as a special case and has been extended to incorporate latent variables~\cite{Jebara:thesis01,Zhu:medlda} and to perform structured output prediction~\cite{Zhu:jmlr09}.
Recent work has further extended MED to unite Bayesian nonparametrics
and max-margin learning~\cite{Zhu:icml11,Zhu:regBayes14}, which have been largely treated as isolated topics, for learning better classification models.
The present work contributes by introducing a novel generalization of MED to perform the challenging task of predicting relational links.

Finally, some preliminary results were reported in~\cite{Zhu:icml12}. This paper presents a systematic extension with an efficient stochastic variational algorithm and the empirical results on various large-scale networks.

\vspace{-.15cm}
\section{Max-margin Latent Feature Models}
\vspace{-.1cm}

We now present our max-margin latent feature relational model with an efficient inference algorithm.

\vspace{-.2cm}
\subsection{Latent Feature Relational Models}


Assume we have an $N \times N$ relational link matrix $Y$, where $N$ is the number of entities. We consider the binary case, where the entry $Y_{ij} = +1$ (or $Y_{ij} = -1$) indicates
the presence (or absence) of a link between entity $i$ and entity $j$. We emphasize that all
the latent feature models introduced below can be extended to deal with real or categorical $Y$.\footnote{For LFRMs, this can be done by defining a proper $\Phi$ function in Eq.~(\ref{Eq:LinkLikelihood}). For MedLFRM, this can be done by defining a proper hinge-loss, similar as in~\cite{Zhu:medlda}.}
We consider the link prediction in static networks, where $Y$ is not fully observed and
the goal of link prediction is to learn a model from observed links such that we can predict
the values of unobserved entries of $Y$. In some cases, we may have observed attributes $X_{ij} \in \mathbb{R}^D$ that
affect the link between $i$ and $j$.

In a latent feature relational model, each entity is associated with a vector $\mu_i \in \mathbb{R}^K$, a point in a latent feature space (or latent social space).
Then, the probability of a link can be generally defined as\vspace{-.2cm}
\begin{eqnarray}\label{Eq:LinkLikelihood}
p(Y_{ij}=1|X_{ij}, \mu_i, \mu_j) = \Phi\big( \psi(\mu_i, \mu_j) + \eta^\top X_{ij} + b \big),\vspace{-.3cm}
\end{eqnarray}
where a common choice of $\Phi$ is the sigmoid function\footnote{Other choices exist, such as the probit function~\cite{Chang:RTM09,Chen:pami15}.}, i.e., $\Phi(t) = \frac{1}{1 + e^{-t}}$; $\psi(\mu_i, \mu_j)$ is a function that measures how similar the two entities $i$ and $j$ are in the latent social space; the observed attributes $X_{ij}$ come into the likelihood under a generalized linear model; and $b$ is an offset.
The formulation in (\ref{Eq:LinkLikelihood}) covers various interesting cases, including (1)
{\it latent distance model}~\cite{Hoff:02}, which defines $\psi(\mu_i, \mu_j) = - d(\mu_i, \mu_j)$,
using a distance function~$d(\cdot)$ in the latent space; and (2)
{\it latent eigenmodel}~\cite{hoff}, which generalizes the latent distance model and the latent class model for modeling symmetric relational data, and defines
$\psi(\mu_i, \mu_j) = \mu_i^\top D \mu_j$, where~$D$ is a diagonal matrix that is estimated from observed data.

In the above models, the dimension $K$ of the latent social space is assumed to be given {\it a priori}. For a given network, a model selection procedure (e.g., cross-validation) is needed to choose a good value. The nonparametric latent feature relational model (LFRM)~\cite{Miller:nips09} leverages the recent advances in Bayesian nonparametrics to automatically infer
the latent dimension from observed data. Specifically, LFRM assumes that each entity is associated with an infinite dimensional binary vector\footnote{Real-valued features are possible, e.g., by element-wisely multiplying a multivariate Gaussian variable. In the infinite case, this actually defines a Gaussian process.} $\mu_i \in \{0, 1\}^\infty$ and define the discriminant function as
\begin{eqnarray}\label{eq:lfrm}
 \psi(\mu_i, \mu_j) = \mu_i^\top W \mu_j, 
\end{eqnarray}
where $W$ is a weight matrix.
We will use $Z$ to denote a binary feature matrix, where each row corresponds to the latent feature of an entity. For LFRM, we have
$Z = [\mu_1^\top;  \cdots; \mu_N^\top ]$. In LFRM, Indian buffet process (IBP)~\cite{Griffiths:nips06}
was used as the prior of $Z$ to induce a sparse latent feature vector for each entity. The nice properties of IBP ensure that for a fixed dataset a finite number of features suffice to fit the data.
Full Bayesian inference 
with MCMC sampling is usually performed for these models by imposing appropriate priors on latent features and model parameters.

Miller et al.~\cite{Miller:nips09} discussed the more flexible expressiveness of LFRM over latent class models. Here, we provide another support for the expressiveness over the latent egienmodel. For modeling symmetric relational data, we usually constrain $W$ to be symmetric~\cite{Miller:nips09}. Since a symmetric real matrix is diagonalizable,
we can find an orthogonal matrix $Q$ satisfying that $Q^\top W Q$ is a diagonal matrix, denoted again by $D$. Therefore, we have $W = Q D Q^\top$.
Plugging the expression into (\ref{eq:lfrm}), we can treat $ZQ$ as the effective real-valued latent features and conclude that {\it LFRM reduces to a latent eigenmodel for modeling symmetric relational data}. But LFRM is more flexible on adopting an asymmetric weight matrix $W$ and allows the number of factors being unbounded.

\vspace{-.1cm}
\subsection{Max-margin Latent Feature Models}
\vspace{-.1cm}

We now present the max-margin nonparametric latent feature model and its variational inference algorithm.

\begin{table}\vspace{-.2cm}
	\centering
	\caption{Major notations used for MedLFRM.} \vspace{-.2cm}
	\begin{tabular}{|l|l|l|l|}
		\hline
		$N, K$ & \multicolumn{3}{l|}{number of entities and number of features} \\ \hline 
		$X_{ij}, Y_{ij}$ & \multicolumn{3}{l|}{observed attributes and link between entities $i$ and $j$} \\\hline
        $Z$ & (binary) feature matrix & $\nu$ & auxiliary variables \\\hline
		$C$ & regularization parameter & $\mathcal{I}$ & set of training links \\\hline
		$W, \eta, \Theta$ & \multicolumn{3}{l|}{feature weights for $Z$ and $X_{ij}$, $\Theta = \{W, \eta\}$} \\\hline
		$\gamma, \psi$ & \multicolumn{3}{l|}{variational parameter for $\nu$ and $Z$}  \\\hline
		$\Lambda, \kappa$ & \multicolumn{3}{l|}{posterior mean of $W$ and $\eta$} \\\hline
	\end{tabular}\vspace{-.3cm}
\end{table}

\vspace{-.1cm}
\subsubsection{MED Latent Feature Relational Model}
\vspace{-.1cm}


We follow the same setup as the general LFRM model, and represent each entity using a set of binary features.
Let $Z$ denote the binary feature matrix,
of which each row corresponds to an entity and each column corresponds to a feature. The entry $Z_{ik}=1$ means that entity $i$ has feature $k$; and $Z_{ik}=0$ denotes that entity $i$ does not has feature $k$. Let $\Theta$ denote the model parameters. We share the same goal as LFRM to learn a posterior distribution $p(\Theta, Z | X, Y)$, but with a fundamentally different procedure, as detailed below.

If the features $Z_i$ and $Z_j$ are given, we 
define the {\it latent discriminant function} as
\setlength\arraycolsep{1pt} \begin{eqnarray}
f(Z_i, Z_j; X_{ij}, W, \eta) &=& Z_i W Z_j^\top + \eta^\top X_{ij} , 
\end{eqnarray}
where $W$ is a real-valued matrix and the observed attributes (if any) again come into play via a linear model with weights $\eta$.
The entry $W_{kk^\prime}$ is the weight that affects
the link from entity $i$ to entity $j$ if entity $i$ has feature $k$ and entity $j$ has feature $k^\prime$.
In this model, we have $\Theta = \{W, \eta\}$.

To perform Bayesian inference, we define a prior $p_0(\Theta, Z) = p_0(\Theta) p_0(Z)$.
For finite sized matrices $Z$ with $K$ columns, we can define the prior $p_0(Z)$ as a Beta-Bernoulli process~\cite{Meeds:nips07}.
In the infinite case, where $Z$ has an infinite number of columns, 
we adopt the Indian buffet process (IBP) prior over the unbounded binary matrices as described in~\cite{Griffiths:nips06}. The prior $p_0(\Theta)$ can be the common Gaussian.

To predict the link between entities $i$ and $j$, we need to get rid of the uncertainty of latent variables. We follow the strategy that has proven effective in various tasks~\cite{Zhu:medlda} and define the {\it effective discriminant function}:\footnote{An alternative strategy is to learn a Gibbs classifier, which can lead to a closed-form posterior allowing MCMC sampling, as discussed in~\cite{Bei-Link:16,Zhu:GibbsMedLDA}. }
\begin{eqnarray}
f(X_{ij}) = \ep_{p(Z, \Theta)}[f(Z_i, Z_j; X_{ij}, \Theta)].
\end{eqnarray}
Then, the prediction rule for binary links is $\hat{Y}_{ij} = \sign f(X_{ij}).$
Let $\mathcal{I}$ denote the set of pairs that have observed links in training set. The training error will be
$\mathcal{R}_{tr} = \sum_{(i,j) \in \mathcal{I}} \ell \indicator( Y_{ij} \neq \hat{Y}_{ij} )$, where $\ell$ is
a positive cost parameter and $\indicator(\cdot)$ is an indicator function that equals $1$ if the predicate holds, otherwise $0$.
Since the training error is hard to deal with due to its non-convexity, we often find a good surrogate loss.
We choose the well-studied hinge-loss, which is convex. In our case, we can show that the following hinge-loss\vspace{-.2cm}
\begin{eqnarray}
\mathcal{R}_{\ell}(p(Z, \Theta)) =  \sum_{(i,j) \in \mathcal{I}} h_{\ell}( Y_{ij} f(X_{ij}) ),\vspace{-.3cm}
\end{eqnarray}
is an upper bound of the training error $\mathcal{R}_{tr}$. 

Then, we define the MED latent feature relational model (MedLFRM) as solving the problem\vspace{-.2cm}
\setlength\arraycolsep{-2pt}\begin{eqnarray}
&&\min_{p(Z, \Theta) \in \mathcal{P}} \KL(p(Z, \Theta) \Vert p_0(Z, \Theta)) + C \cdot \mathcal{R}_\ell( p(Z, \Theta) ),  \vspace{-.3cm} 
\end{eqnarray}
where $C$ is a positive regularization parameter balancing the influence between the prior and the large-margin hinge-loss; and $\mathcal{P}$ denotes the space of normalized distributions.

For the IBP prior, it is often more convenient to deal with the stick-breaking representation~\citep{YWTeh:aistats07}, which introduces some auxiliary variables and converts marginal dependencies into conditional independence.
Specifically, let $\pi_k \in (0, 1)$ be a parameter associated with column $k$ of $Z$.
The parameters $\piv$ are generated by a stick-breaking process, that is,\vspace{-.2cm}
\setlength\arraycolsep{1pt}\begin{eqnarray}
\forall i:~ \nu_i &\sim& \B(\alpha, 1), \nonumber \\
\forall k:~ \pi_k &=& \nu_k \pi_{k-1} = \prod_{i=1}^k \nu_i,~\textrm{where}~\pi_0=1. \nonumber \vspace{-.3cm}
\end{eqnarray}
Given $\pi_k$, each $Z_{nk}$ in column $k$ is sampled independently from $\Ber(\pi_k)$.
This process results in a decreasing sequence of probabilities $\pi_k$, and
the probability of seeing feature $k$ decreases exponentially with $k$ on a finite dataset.
In expectation, only a finite of features will be active for a given finite dataset.
With this representation, we have the augmented MedLFRM\vspace{-.2cm}
\setlength\arraycolsep{-5pt} \begin{eqnarray}\label{problem:augMedLFRM}
&& \min_{p(\nuv, Z, \Theta) } \KL(p(\nuv, Z, \Theta) \Vert p_0(\nuv, Z, \Theta)) + C \cdot \mathcal{R}_\ell( p(Z, \Theta) ),  \vspace{-.3cm} 
\end{eqnarray}
where the prior has a factorization form $p_0(\nuv, Z, \Theta) = p_0(\nuv)p(Z|\nuv)p_0(\Theta)$.

We make several comments about the above definitions. First, we have adopted the similar method as in~\cite{Zhu:icml11,Zhu:regBayes14}
to define the discriminant function using the expectation operator, instead of the traditional
log-likelihood ratio of a Bayesian generative model with latent variables~\cite{Jebara:thesis01,Lewis:06}. The linearity of expectation
makes our formulation simpler than the one that could be achieved by using a highly nonlinear log-likelihood ratio.
Second, although a likelihood model can be defined as in~\cite{Zhu:icml11,Zhu:regBayes14} to perform hybrid learning,
we have avoided doing that because the sigmoid link likelihood model in Eq. (\ref{Eq:LinkLikelihood}) is
highly nonlinear and it could make the hybrid problem hard to solve. Finally, though the target distribution is
the augmented posterior $p(\nuv, Z, \Theta)$, the hinge loss only depends on the marginal
distribution $p(Z, \Theta)$, with the augmented variables $\nuv$ collapsed out. This does not cause any
inconsistency because the effective discriminative function $f(X_{ij})$ (and thus the hinge loss) only depends
on the marginal distribution with $\nuv$ integrated out even if we take the expectation
with respect to the augmented posterior.

\vspace{-.1cm}
\subsubsection{Inference with Truncated Mean-Field}

We now present a variational algorithm for posterior inference. In next section, we will present
a more efficient extension by doing stochastic subsampling.

We note that problem~(\ref{problem:augMedLFRM}) has nice properties. For example, the hinge loss
$\mathcal{R}_\ell$ is a piece-wise linear functional of $p$ and the discriminant function $f$ is linear of the weights $\Theta$.
While sampling methods could lead to more accurate results, variational methods~\cite{Jordan:99} are
usually more efficient and they also have an objective to monitor the convergence behavior.
Here, we introduce a simple variational method to explore such properties, which turns out to
perform well in practice.
Specifically, let $K$ be a truncation level. We make the truncated mean-field assumption\vspace{-.2cm}
\setlength\arraycolsep{-2pt}\begin{eqnarray}
&&p(\nuv, Z, \Theta) = p(\Theta) \prod_{k=1}^K p(\nu_k | \gamma_k) \left( \prod_{i=1}^N p(Z_{ik} | \psi_{ik}) \right),\vspace{-.3cm}
\end{eqnarray}
where $p(\nu_k | \gamma_k) = \B(\gamma_{k1}, \gamma_{k2})$, $p(Z_{ik} | \psi_{ik}) = \Ber(\psi_{ik})$ are the variational
distributions with parameters $\{\gamma_k, \psi_{ik}\}$. Note that the truncation error of marginal distributions decreases exponentially as $K$ increases~\cite{YWTeh:aistats09}. In practice, a reasonably large $K$ will be sufficient as shown in experiments.
Then, we can solve problem (\ref{problem:augMedLFRM}) with an iterative procedure that alternates between:

{\bf Solving for $p(\Theta)$}: by fixing $p(\nuv, Z)$ and ignoring irrelevant terms, the subproblem can be
equivalently written in a constrained form\vspace{-.2cm}
\setlength\arraycolsep{1pt} \begin{eqnarray}
\min_{p(\Theta), \xiv} && \KL(p(\Theta) \Vert p_0(\Theta)) + C \sum_{(i,j) \in \mathcal{I}} \xi_{ij} \\
\forall (i,j) \in \mathcal{I}, ~ \mathrm{s.t.:} && Y_{ij} (\mathrm{Tr}(\ep[W] \bar{\Zv}_{ij}) + \ep[\eta]^\top X_{ij}) \geq \ell - \xi_{ij}, \nonumber \vspace{-.3cm}
\end{eqnarray}
where $\bar{\Zv}_{ij} = \ep_{p}[Z_j^\top Z_i]$ is the expected latent features under the current distribution $p(Z)$, $\textrm{Tr}(\cdot)$ is the trace of a matrix, and $\xiv =\{\xi_{ij}\}$ are slack variables.
By Lagrangian duality theory, we have the optimal solution\vspace{-.2cm}
\setlength\arraycolsep{1pt}
\begin{eqnarray}
p(\Theta) \propto p_0(\Theta) \exp \Big\{ \sum_{(i,j) \in \mathcal{I}} \omega_{ij} Y_{ij} ( \mathrm{Tr}(W \bar{\Zv}_{ij}) + \eta^\top X_{ij}) \Big\}, \nonumber \vspace{-.3cm}
\end{eqnarray}
where $\omegav = \{\omega_{ij}\}$ are Lagrangian multipliers.

For the commonly used standard normal prior $p_0(\Theta)$, we have the optimal solution\vspace{-.2cm}
\begin{eqnarray}
p(\Theta) = p(W) p(\eta) = \Big( \prod_{kk^\prime} \mathcal{N}(\Lambda_{kk^\prime}, 1) \Big) \Big( \prod_d \mathcal{N}(\kappa_d , 1) \Big), \nonumber \vspace{-.3cm}
\end{eqnarray}
where the means are $\Lambda_{kk^\prime} = \sum_{(i,j) \in \mathcal{I}}\omega_{ij}Y_{ij} \ep[Z_{ik} Z_{jk^\prime}]$, and $\kappa_d = \sum_{(i,j) \in \mathcal{I}} \omega_{ij} Y_{ij} X_{ij}^d.$ The dual problem is\vspace{-.2cm}
\setlength\arraycolsep{1pt} \begin{eqnarray}
\max_{\omegav} ~&& \ell \sum_{(i,j) \in \mathcal{I}}  \omega_{ij} - \frac{1}{2} (\Vert \Lambda \Vert_2^2 + \Vert \kappa \Vert_2^2) \nonumber \\
\mathrm{s.t.:} ~&& 0 \leq \omega_{ij} \leq C,~\forall (i,j) \in \mathcal{I}. \nonumber
\end{eqnarray}
Equivalently, the mean parameters $\Lambda$ and $\kappa$ can be directly obtained by solving the primal problem\vspace{-.2cm}
\setlength\arraycolsep{1pt} \begin{eqnarray}\label{eq:PrimalSVMNormal}
\min_{\Lambda, \kappa, \xiv} && \frac{1}{2} (\Vert \Lambda \Vert_2^2 + \Vert \kappa \Vert_2^2) + C \sum_{(i,j) \in \mathcal{I}} \xi_{ij} \\
\forall (i,j) \in \mathcal{I}, \mathrm{s.t.:} && Y_{ij} (\mathrm{Tr}(\Lambda \bar{\Zv}_{ij}) + \kappa^\top X_{ij}) \geq \ell - \xi_{ij}, \nonumber \vspace{-.35cm}
\end{eqnarray}
which is a binary classification SVM. We can solve it with any existing high-performance solvers, such as SVMLight or LibSVM.

{\bf Solving for $p(\nuv, Z)$}: by fixing $p(\Theta)$ and ignoring irrelevant terms, 
the subproblem involves solving\vspace{-.2cm}
\begin{eqnarray}
\min_{p(\nuv, Z)} \KL(p(\nuv, Z) \Vert p_0(\nuv, Z)) + C \cdot \mathcal{R}_\ell(p(Z, \Theta)). \nonumber  \vspace{-.35cm} 
\end{eqnarray}
With the truncated mean-field assumption, we have\vspace{-.2cm}
\setlength\arraycolsep{1pt} \begin{eqnarray}
\mathrm{Tr}(\Lambda \bar{\Zv}_{ij} ) = \left\{\begin{array}{ll}
\psi_i \Lambda \psi_j^\top   & \textrm{if}~ i \neq j \\  
\psi_i \Lambda \psi_i^\top + \sum_k \Lambda_{kk} \psi_{ik}(1-\psi_{ik})  & \textrm{if}~ i = j  
\end{array} \right. \nonumber \vspace{-.35cm}
\end{eqnarray}
We defer the evaluation of the KL-divergence to Appendix A. For $p(\nuv)$, since the margin
constraints are not dependent on $\nuv$, we can get the same solutions as in~\cite{YWTeh:aistats09}. Below, we focus on solving for $p(Z)$.

Specifically, we can solve for $p(Z)$ using sub-gradient methods. Define\vspace{-.2cm}
\setlength\arraycolsep{1pt} \begin{eqnarray}
\mathcal{I}_i &&= \{j: j \neq i,~(i,j) \in \mathcal{I}~ \textrm{and}~Y_{ij} f(X_{ij}) \leq \ell \}  \nonumber \\ 
\mathcal{I}_i^\prime &&= \{j: j \neq i,~(j, i) \in \mathcal{I}~ \textrm{and}~Y_{ji} f(X_{ji}) \leq \ell \}. \nonumber \vspace{-.35cm} 
\end{eqnarray}
Intuitively, we can see that $\mathcal{I}_i$ denotes the set of out-links of entity $i$ in the training set, for which the current model has a low confidence on accurate predictions, while $\mathcal{I}^\prime_i$ denotes the set of in-links of entity $i$ in the training set, for which the current model does not have a high confidence on making accurate prediction. For undirected networks, the two sets are identical and we should have only one of them.

Due to the fact that $\partial_x h_\ell(g(x))$ equals to $- \partial_x g(x)$ if $g(x) \leq \ell$; $0$ otherwise, we have the subgradient\vspace{-.2cm}
\setlength\arraycolsep{1pt} \begin{eqnarray}
\partial_{\psi_{ik}} \mathcal{R}_\ell = && - \sum_{j \in \mathcal{I}_i} Y_{ij} \Lambda_{k \cdot} \psi_{j}^\top -
\sum_{j \in \mathcal{I}_i^\prime} Y_{ji} \psi_{j} \Lambda_{\cdot k}  \nonumber \\
&& - \indicator(Y_{ii} f(X_{ii}) \leq \ell) Y_{ii}(\Lambda_{kk}(1-\psi_{ik}) + \Lambda_{k \cdot} \psi_{i}^\top), \nonumber  \vspace{-.35cm}  
\label{eq:subgrad}
\end{eqnarray}
where $\Lambda_{k\cdot}$ denotes the $k$th row of $\Lambda$ and $\Lambda_{\cdot k}$ denotes the $k$th column of $\Lambda$.
Note that for the cases where we do not consider the self-links, the third term will not present. 
Moreover, $\partial_{\psi_{ik}} \mathcal{R}_\ell$ does not depend on $\psi_{ik}$.
Then, let the subgradient equal to $0$, and we get the update equation\vspace{-.2cm}
\setlength\arraycolsep{1pt} \begin{eqnarray}
\psi_{ik} = \Phi\left( \sum_{j=1}^k \ep_p[\log \nu_j] - \mathcal{L}_k^\nu - C \cdot \partial_{\psi_{ik}} \mathcal{R}_\ell \right),
\label{eq:phi} \vspace{-.35cm}
\end{eqnarray}
where $\mathcal{L}_k^\nu$ is a lower bound of $\ep_p[\log(1 - \prod_{j=1}^k \nu_j)]$. For clarity, we defer the details to Appendix A.

\vspace{-.1cm}
\subsubsection{Stochastic Variational Inference}
The above batch algorithm needs to scan the full training set at each iteration, which can be prohibitive for large-scale networks. When $W$ is a full matrix, at each iteration the complexity of updating $p(\nuv, Z)$ is $\mathcal{O}( (N + |\mathcal{I}|) K^2 )$, while the complexity of computing $p(\Theta)$ is $\mathcal{O}( |\mathcal{I}| K^2)$ for linear SVMs, thanks to the existing high-performance solvers for linear SVMs, such as the cutting plane algorithm~\cite{Joachims:06}. Even faster algorithms exist to learn linear SVMs, such as Pegasos~\cite{shalev2011pegasos}, a stochastic gradient descent method that achieves $\epsilon$-accurate solution with $\tilde{O}(1/\epsilon)$ iterations, and the dual coordinate descent method~\cite{hsieh2008dual} which needs $O( \log (1/\epsilon) )$ iterations to get an $\epsilon$-accurate solution.
We empirically observed that the time of solving $p(\nuv, Z)$ is much larger than that of $p(\Theta)$ in our experiments (See Table \ref{table:time-svm-vi}).
Inspired by the recent advances on stochastic variational inference (SVI)~\cite{JMLR:v14:hoffman13a}, we present a stochastic version of our variational method to efficiently handle large networks by random subsampling, as outlined in Alg.~\ref{alg:svi} and detailed below.

The basic idea of SVI is to construct an unbiased estimate of the objective and its gradient. Specifically, under the same mean-field assumption as above, an unbiased estimate of our objective (\ref{problem:augMedLFRM}) at iteration $t$ is
\begin{eqnarray}
\hat{\mathcal{L}}(p(\nuv, Z,\Theta)) &\triangleq& \KL(p(\nuv \Vert \gammav) \Vert p_0(\nuv)) + \KL(p(\Theta) \Vert p_0(\Theta)) \nonumber \\
&& + \frac{N}{|\mathcal{N}_t|} \sum_{i \in \mathcal{N}_t} \ep_p[ \KL(p(Z_i|\psiv_i) \Vert p_0(Z_i | \nuv)) ] \nonumber \\
&& + \frac{|\mathcal{I}|}{|\mathcal{E}_t|} C \sum_{(i,j)\in \mathcal{E}_t} h_\ell( Y_{ij} f(X_{ij}) ) , \nonumber
\end{eqnarray}
where $\mathcal{N}_t$ is the subset of randomly sampled entities; $\mathcal{E}_t$ is the subset of randomly sampled edges\footnote{Sampling a single entity and a single edge at each iteration does not lose the unbiasedness, but it often has a large variance to get unstable estimates. We consider the strategy that uses a mini-batch of entities and a mini-batch of edges to reduce variance.}; and the KL-divergence terms can be evaluated as detailed in Appendix A. There are various choices on drawing samples to derive an unbiased estimate~\cite{gopalan2013efficient}. We consider the simple scheme that first uniformly draws the entities and then uniformly draws the edges associated with the entities in $\mathcal{N}_t$.

\begin{algorithm}[t]
\caption{Stochastic Variational Inference }\label{alg:svi}
\centering
\begin{algorithmic}[1]
    \STATE \textbf{Inputs}: $\kappa_{\gamma}$, $\kappa_{\psi}$, $\mu_{\gamma}$, $\mu_{\psi}$,~ $t=1$
	\REPEAT
    \STATE Select a batch $\mathcal{N}_t$ of entities, and set $\mathcal{E}_t = \emptyset$
    \FORALL{entity $i \in \mathcal{N}_t$}
    	\STATE Select a batch $\mathcal E_t^i$ of links connected to $i$
        \STATE Set $\mathcal{E}_t = \mathcal E_t \cup \mathcal{E}_t^i$
    \ENDFOR
    \FORALL{$k = 1, \dots, K$}
    \STATE Obtain $\hat{\gamma}_k$ by minimizing $\hat{\mathcal{L}}$ similar as in~\cite{YWTeh:aistats09}
    \STATE Set $\rho_t^\gamma = (\mu_\gamma + t)^{-\kappa_{\gamma}}$
    \STATE Set $\gamma_k = (1 - \rho_t^\gamma)\gamma_{k} + \rho_t^\gamma \hat{\gamma}_{k}$
    \FORALL{entity $i \in \mathcal{N}_t$}
        \STATE Obtain $\hat{\psi}_{ik}$ according to Eq.~(\ref{eq:subgradnew})
        \STATE Set $\rho_{t,\psi} = (\mu_\psi + t)^{-\kappa_{\psi}}$
       	\STATE Set $\psi_{ik} = (1-\rho_{t,\psi})\psi_{ik} + \rho_{t,\psi}\hat{\psi}_{ik}$
    \ENDFOR
    \ENDFOR
    \STATE Update $p(\Theta)$ using the subset $\mathcal{E}_t$ of edges
    \STATE Set $t = t + 1$
    \UNTIL{$\gamma$ and $\psi$ are optimal}
\end{algorithmic}
\end{algorithm}

Then, we can follow the similar procedure as in the batch algorithm to optimize $\hat{\mathcal{L}}$ to solve for $p(\nuv, Z)$ and $p(\Theta)$.
For $p(\nuv)$, the update rule is almost the same as in the batch algorithm, except that we only need to consider the subset of entities in $\mathcal{N}_t$ with a scaling factor of $\frac{N}{|\mathcal{N}_t|}$. For $p(Z)$, due to the mean-field assumption, we only need to compute $p(Z_i)$ where $i \in \mathcal{N}_t$ at iteration $t$. Let $\hat{\mathcal{R}}_t \triangleq \sum_{(i,j)\in \mathcal{E}_t} h_\ell( Y_{ij} f(X_{ij}) ) $. For each $p(Z_i)$, we can derive the (unbiased) stochastic subgradient:
\setlength\arraycolsep{1pt} \begin{eqnarray}
\partial_{\psi_{ik}} \hat{\mathcal{R}}_t = && - \sum_{j \in \mathcal{E}_{it} } Y_{ij} \Lambda_{k \cdot} \psi_{j}^\top -
\sum_{j \in \mathcal{E}_{it}^\prime} Y_{ji} \psi_{j} \Lambda_{\cdot k}  \nonumber \\
&& - \indicator(i \in \mathcal{N}_t) Y_{ii}(\Lambda_{kk}(1-\psi_{ik}) + \Lambda_{k \cdot} \psi_{i}^\top), \nonumber   
\label{eq:subgradnew}
\end{eqnarray}
where the two subsets are defined as
\setlength\arraycolsep{1pt} \begin{eqnarray}
\mathcal{E}_{it} &&= \{j: j \neq i,~(i,j) \in \mathcal{E}_t  ~ \textrm{and}~Y_{ij} f(X_{ij}) \leq \ell \}  \nonumber \\ 
\mathcal{E}_{it}^\prime &&= \{j: j \neq i,~(j, i) \in \mathcal{E}_t ~ \textrm{and}~Y_{ij} f(X_{ij}) \leq \ell \}. \nonumber 
\end{eqnarray}
Setting the subgradient at zero leads to the closed-form update rule:
\setlength\arraycolsep{-3pt} \begin{eqnarray}
&&\hat{\psi}_{ik} = \Phi\left( \sum_{j=1}^k \ep_p[\log \nu_j] - \mathcal{L}_k^\nu - \frac{|\mathcal{I}|}{ |\mathcal{E}_t| } C \cdot \partial_{\psi_{ik}} \hat{\mathcal{R}}_t \right).
\label{eq:phi}
\end{eqnarray}
Finally, the substep of updating $p(\Theta)$ still involves solving an SVM problem, which only needs to consider the sampled edges in the set $\mathcal{E}_t$, a much smaller problem than the original SVM problem that handles $|\mathcal{I}|$ number of edges.

We optimize the unbiased objective $\hat{\mathcal{L}}$ by specifying a learning rate $\rho_t = (\mu + t)^{-\kappa}$ at iteration $t$, which is similar to \cite{JMLR:v14:hoffman13a}. However, different from \cite{JMLR:v14:hoffman13a}, we select values of $\kappa$ between 0 and 1. For $\kappa \in [0, 0.5]$, this breaks the local optimum convergence conditions of the Robbins-Monro algorithm, but allows for larger update steps at each iteration. Empirically, we can arrive at a satisfying solution faster using $\kappa \in [0, 0.5]$. We use different $\kappa$'s when updating $p(\nu)$ and $p(Z)$, which we denote as $\kappa_\gamma$ and $\kappa_\psi$.

\vspace{-.15cm}
\subsection{The Full Bayesian Model}

MedLFRM has a regularization parameter $C$, which normally plays an important role
in large-margin classifiers, especially on sparse and imbalanced datasets.
To search for a good $C$, cross-validation is a typical approach, but
it could be computationally expensive by comparing many candidates. Under the probabilistic
formulation, we provide a full Bayesian formulation of MedLFRM, which avoids hyper-parameter tuning.
Specifically, if we divide the objective by $C$, the KL-divergence term will have an extra parameter $1/C$.
Below, we present a hierarchical prior to avoid explicit tuning of regularization parameters, which essentially infers $C$ as detailed after Eq.~(\ref{eq:PrimalSVMGammaNormal}).


{\bf Normal-Gamma Prior}: For simplicity, we assume that the prior on $\Theta$ is an isotropic normal distribution\footnote{A more flexible prior will be the one that uses different means and variances for different components of $\Theta$. 
} with common mean $\mu$ and precision $\tau$\vspace{-.2cm}
\begin{eqnarray}
p_0(\Theta | \mu, \tau) = \prod_{k k^\prime} \mathcal{N}(\mu, \tau^{-1}) \prod_d \mathcal{N}(\mu, \tau^{-1}).\vspace{-.35cm}
\end{eqnarray}
We further use a Normal-Gamma hyper-prior for $\mu$ and $\tau$, denoted by $\mathcal{NG}(\mu_0, n_0, \frac{\nu_0}{2}, \frac{2}{S_0})$:\vspace{-.2cm}
\begin{eqnarray}
p_0(\mu|\tau) = \mathcal{N}(\mu_0, (n_0 \tau)^{-1}),~ p_0(\tau) = \mathcal{G}(\frac{\nu_0}{2}, \frac{2}{S_0}), \nonumber \vspace{-.35cm}
\end{eqnarray}
where $\mathcal{G}$ is the Gamma distribution, $\mu_0$ is the prior mean, $\nu_0$ is the prior degree of freedom, $n_0$ is the prior sample size,
and $S_0$ is the prior sum of squared errors.

We note that the normal-Gamma prior has been used in a marginalized form as a heavy-tailed prior for deriving sparse estimates~\cite{Griffin:2010}.
Here, we use it for automatically inferring the regularization constants, which replace the role of $C$ in problem~(\ref{problem:augMedLFRM}).
Also, our Bayesian approach is different from the previous methods for
estimating the hyper-parameters of SVM, by optimizing a log-evidence~\cite{Gold:nn05} or an estimate of the generalization error~\cite{Chapelle:ml02}.

Formally, with the above hierarchical prior, we define Bayesian MedLFRM (BayesMedLFRM) as solving\vspace{-.2cm}
\setlength\arraycolsep{-2.5pt} \begin{eqnarray}\label{problem:augBayesMedLFRM}
&& \min_{p(\nuv, Z, \mu, \tau, \Theta)} \left\{ \begin{array}{c}
 \KL(p(\nuv, Z, \mu, \tau, \Theta) \Vert p_0(\nuv, Z, \mu, \tau, \Theta)) \\
+ \mathcal{R}_\ell( p(Z, \Theta) )  \end{array} \right\}, \vspace{-0.35cm} 
\end{eqnarray}
where $p_0(\nuv, Z, \mu, \tau, \Theta) \!=\! p_0(\nuv, Z) p_0(\mu, \tau) p_0(\Theta|\mu, \tau)$.
For this problem, we can develop a similar iterative algorithm as for MedLFRM, where the sub-step of inferring $p(\nuv, Z)$ does not change.
For $p(\mu, \tau, \Theta)$, by introducing slack variables the sub-problem can be equivalently written in a constrained form:\vspace{-.2cm}
\setlength\arraycolsep{1pt}\begin{eqnarray}\label{eq:VarNormalGamma}
\min_{p(\mu, \tau, \Theta), \xiv} \KL( &&p(\mu, \tau, \Theta) \Vert p_0(\mu, \tau, \Theta)) + \sum_{(i,j) \in \mathcal{I}} \xi_{ij}  \\
\forall (i,j) \in \mathcal{I}, \mathrm{s.t.:} && Y_{ij} (\mathrm{Tr}(\ep[W] \bar{\Zv}_{ij}) + \ep[\eta]^\top X_{ij}) \geq \ell - \xi_{ij}, \nonumber \vspace{-.35cm}
\end{eqnarray}
which is convex but intractable to solve directly. Here, we make the mild mean-field assumption that $p(\mu, \tau, \Theta) = p(\mu, \tau) p(\Theta)$.
Then, we iteratively solve for $p(\Theta)$ and $p(\mu, \tau)$, as summarized below. We defer the details to Appendix B.

For $p(\Theta)$, we have the mean-field update equation\vspace{-.2cm}
\begin{eqnarray}
p(W_{kk^\prime}) = \mathcal{N}(\Lambda_{kk^\prime}, \lambda^{-1} ),~p(\eta_d) = \mathcal{N}(\kappa_d, \lambda^{-1}), \vspace{-.35cm}
\end{eqnarray}
where $\Lambda_{kk^\prime} \!=\! \ep[\mu] \!+\! \lambda^{-1} \sum_{(i,j) \in \mathcal{I}} \omega_{ij}Y_{ij} \ep[Z_{ik}Z_{jk^\prime}]$, $\kappa_d \!=\! \ep[\mu] + \lambda^{-1} \sum_{(i,j) \in \mathcal{I}} \omega_{ij}Y_{ij} X^d_{ij},$ and $\lambda \!=\! \ep[ \tau ]$.
Similar as in MedLFRM, the mean of $\Theta$ can be obtained by solving the following problem \vspace{-.2cm}
\begin{eqnarray}
\min_{\Lambda, \kappa, \xiv} && ~\frac{\lambda}{2}( \Vert \Lambda - \ep[\mu] E \Vert_2^2 + \Vert \kappa - \ep[\mu] \ev \Vert_2^2) + \sum_{(i,j) \in \mathcal{I}} \xi_{ij} \nonumber \\
\mathrm{s.t.:} && ~ Y_{ij} (\mathrm{Tr}(\Lambda \bar{\Zv}_{ij}) + \kappa^\top X_{ij}) \geq \ell - \xi_{ij},~\forall (i,j) \in \mathcal{I}, \nonumber  \vspace{-.35cm}
\end{eqnarray}
where $\ev$ is a $K \times 1$ vector with all entries being the unit 1 and $E = \ev \ev^\top$ is a $K \times K$ matrix.
Let $\Lambda^\prime = \Lambda - \ep[\mu] E$ and $\kappa^\prime = \kappa - \ep[\mu] \ev$, we have the transformed problem\vspace{-.2cm}
\begin{eqnarray}\label{eq:PrimalSVMGammaNormal}
\min_{\Lambda^\prime, \kappa^\prime, \xiv} && \frac{\lambda}{2} (\Vert \Lambda^\prime \Vert_2^2 + \Vert \kappa^\prime \Vert_2^2) + \sum_{(i,j) \in \mathcal{I}} \xi_{ij}  \\
\forall (i,j) \in \mathcal{I},~ \mathrm{s.t.:} && Y_{ij} (\mathrm{Tr}(\Lambda^\prime \bar{\Zv}_{ij}) + (\kappa^\prime)^\top X_{ij}) \geq \ell_{ij} - \xi_{ij} \nonumber \vspace{-.35cm}
\end{eqnarray}
where $\ell_{ij} = \ell - \ep[\mu] Y_{ij}(\mathrm{Tr}(E\bar{\Zv}_{ij}) + \ev^\top X_{ij})$ is the adaptive cost.
The problem can be solved using an existing binary SVM solver with slight changes to consider the sample-varying costs.
Comparing with problem~(\ref{eq:PrimalSVMNormal}), we can see that BayesMedLFRM automatically infers the
regularization constant $\lambda$ (or equivalently $C$), by iteratively updating the posterior distribution $p(\tau)$, as explained below.

\begin{table*}[t]\vspace{-.2cm}
\caption{AUC on the countries and kinship datasets. Bold indicates the best performance.}
\label{table:nation}\vspace{-.4cm}
\begin{center}
\begin{tabular}{|c|c|c|c|c|}
\hline
{} & Countries (single) & Countries (global) & Kinship (single) & Kinship (global) \\
\hline
SVM         & 0.8180 $\pm$ 0.0000 & 0.8180 $\pm$ 0.0000  & --- & --- \\
LR          & 0.8139 $\pm$ 0.0000 & 0.8139 $\pm$ 0.0000  & --- & --- \\
\hline
MMSB        & 0.8212 $\pm$ 0.0032 & 0.8643 $\pm$ 0.0077  & 0.9005 $\pm$ 0.0022 & 0.9143 $\pm$ 0.0097 \\
IRM		    & 0.8423 $\pm$ 0.0034 & 0.8500 $\pm$ 0.0033  & 0.9310 $\pm$ 0.0023 & 0.8943 $\pm$ 0.3000 \\
LFRM rand   & 0.8529 $\pm$ 0.0037 & 0.7067 $\pm$ 0.0534  & 0.9443 $\pm$ 0.0018 & 0.7127 $\pm$ 0.0300 \\
LFRM w/ IRM & 0.8521 $\pm$ 0.0035 & 0.8772 $\pm$ 0.0075  & 0.9346 $\pm$ 0.0013 & 0.9183 $\pm$ 0.0108 \\
\hline
MedLFRM	     & {\bf 0.9173} $\pm$ 0.0067 & {\bf 0.9255} $\pm$ 0.0076  & {\bf 0.9552} $\pm$ 0.0065 & {\bf 0.9616} $\pm$ 0.0045 \\
BayesMedLFRM & {\bf 0.9178} $\pm$ 0.0045 & {\bf 0.9260} $\pm$ 0.0023  & {\bf 0.9547} $\pm$ 0.0028 & {\bf 0.9600} $\pm$ 0.0016 \\
\hline
\end{tabular}\vspace{-.6cm}
\end{center}
\end{table*}

The mean-field update equation for $p(\mu, \tau)$ is \vspace{-.2cm}
\begin{eqnarray}
p(\mu, \tau) = \mathcal{NG}( \tilde{\mu}, \tilde{n}, \tilde{\nu}, \tilde{S}), \vspace{-.35cm}
\end{eqnarray}
where $\tilde{\mu} = \frac{K^2 \bar{\Lambda} + D \bar{\kappa} + n_0 \mu_0}{K^2 + D + n_0},$ $\tilde{n} = n_0 + K^2 + D,$ $\tilde{\nu} = \nu_0 + K^2 + D,$
$\tilde{S} = \ep [S_{W}] + \ep [S_{\eta}] + S_0 + \frac{n_0 (K^2 (\bar{\Lambda} - \mu)^2 + D (\bar{\kappa} - \mu)^2) }{K^2 + D + n_0}$,
and $S_W = \Vert W - \bar{W} E\Vert_2^2$,~$S_\eta = \Vert \eta - \bar{\eta} \ev \Vert_2^2$.
From $p(\mu, \tau)$, we can compute the expectation and variance, which are needed in updating $p(\Theta)$ \vspace{-.2cm} 
\begin{eqnarray}\label{eq:postmean-NG}
\ep[ \mu ] = \tilde{\mu},~\ep[ \tau ] = \frac{\tilde{\nu}}{\tilde{S}},~\mathrm{and}~\mathrm{Var}(\mu) = \frac{\tilde{S}}{\tilde{n}(\tilde{\nu} - 2)}. \vspace{-.35cm}
\end{eqnarray}

Finally, similar as in MedLFRM we can develop a stochastic version of the above variational inference algorithm for the full Bayesian model by randomly drawing a mini-batch of entities and a mini-batch of edges at each iteration. The only difference is that we need an extra step to update $p(\mu,\tau)$, which remains the same as in the batch algorithm because both $\mu$ and $\tau$ are global variables shared across the entire dataset.

\vspace{-.1cm}
\section{Experiments}

We provide extensive empirical studies on various real datasets to demonstrate the effectiveness of the max-margin principle in learning
latent feature relational models, as well as the effectiveness of full Bayesian methods in inferring the hyper-parameter $C$.
We also demonstrate the efficiency of our stochastic algorithms on large-scale networks, including the massive US Patents network with millions of nodes.

\vspace{-.1cm}
\subsection{Results with Batch Algorithms}

We first present the results with the batch variational algorithm on relatively small-scale networks.

\vspace{-.1cm}
\subsubsection{Multi-relational Datasets}\label{sec:multirelation}

We report the results of MedLFRM and BayesMedLFRM on the two datasets which were used in~\cite{Miller:nips09}
to evaluate the performance of latent feature relational models. One dataset contains 54 relations of 14 countries
along with 90 given features of the countries, and the other one contains 26 kinship relationships of 104 people
in the Alyawarra tribe in Central Australia. On average, there is a probability of about $0.21$ that a link exists
for each relation on the countries dataset, and the probability of a link is about $0.04$ for the kinship dataset.
So, the kinship dataset is extremely imbalanced (i.e., much more negative examples than positive examples).
To deal with this imbalance in learning MedLFRM, we use different regularization constants
for the positive ($C^+$) and negative ($C^-$) examples. We refer the readers to~\cite{Akbani:ecml04} for other possible choices.
In our experiments, we set $C^+ = 10 C^- = 10 C$ for simplicity and tune the parameter $C$.
For BayesMedLFRM, this equality is held during all iterations, that is, the cost of making an error on positive links is 10 times larger than that on negative links.

Depending on the input data, the latent features might not have interpretable meanings~\cite{Miller:nips09}.
In the experiments, we focus on the effectiveness of max-margin learning
in learning latent feature relational models.
We also compare with two well-established class-based algorithms---IRM
~\cite{Kemp:aaai06}
and MMSB
~\cite{Airoldi:nips08}, both of which were tested in~\cite{Miller:nips09}.
In order to compare with their reported results, we use the same setup for the experiments.
Specifically, for each dataset, we held out $20\%$ of the data during training and
report the AUC (i.e., area under the Receiver Operating Characteristic or ROC curve) for the held out data.
As in~\cite{Miller:nips09}, we consider two settings: (1) ``global" --- we infer a single set of latent features for all relations; and (2) ``single" --- we infer independent latent features for each relation. The overall AUC is
an average of the AUC scores of all relations.

For MedLFRM and BayesMedLFRM, we randomly initialize the posterior mean
of $W$ uniformly in the interval $[0, 0.1]$; initialize $\psi$ to
uniform (i.e., $0.5$) corrupted by a random noise uniformly distributed at $[0, 0.001]$;
and initialize the mean of $\eta$ to be zero.
All the following results of MedLFRM and BayesMedLFRM are averages over 5 randomly initialized runs,
the same as in~\cite{Miller:nips09}. For MedLFRM, the hyper-parameter $C$ is
selected via cross-validation during training. For BayesMedLFRM, we use a very weak hyper-prior with $\mu_0=0$, $n_0=1$, $\nu_0=2$, and $S_0=1$.
We set the cost parameter $\ell = 9$ in all experiments.

Table~\ref{table:nation} shows the results. We can see that in both settings and on both datasets, the max-margin based latent feature relational model
MedLFRM significantly outperforms LFRM that uses a likelihood-based approach with MCMC sampling.
Comparing BayesMedLFRM and MedLFRM, we can see that using the fully-Bayesian technique with a simple Normal-Gamma hierarchical prior,
we can avoid tuning the regularization constant $C$, without sacrificing the link prediction performance.
To see the effectiveness of latent feature models,
we also report the performance of logistic regression (LR) and linear SVM on the countries dataset, which has input features.
We can see that a latent feature or latent class model generally outperforms the methods that are built on raw input features for this particular dataset.

\begin{wrapfigure}{r}{0.24\textwidth}\vspace{-.35cm}
\centering
\includegraphics[height=1in, width=0.24\textwidth]{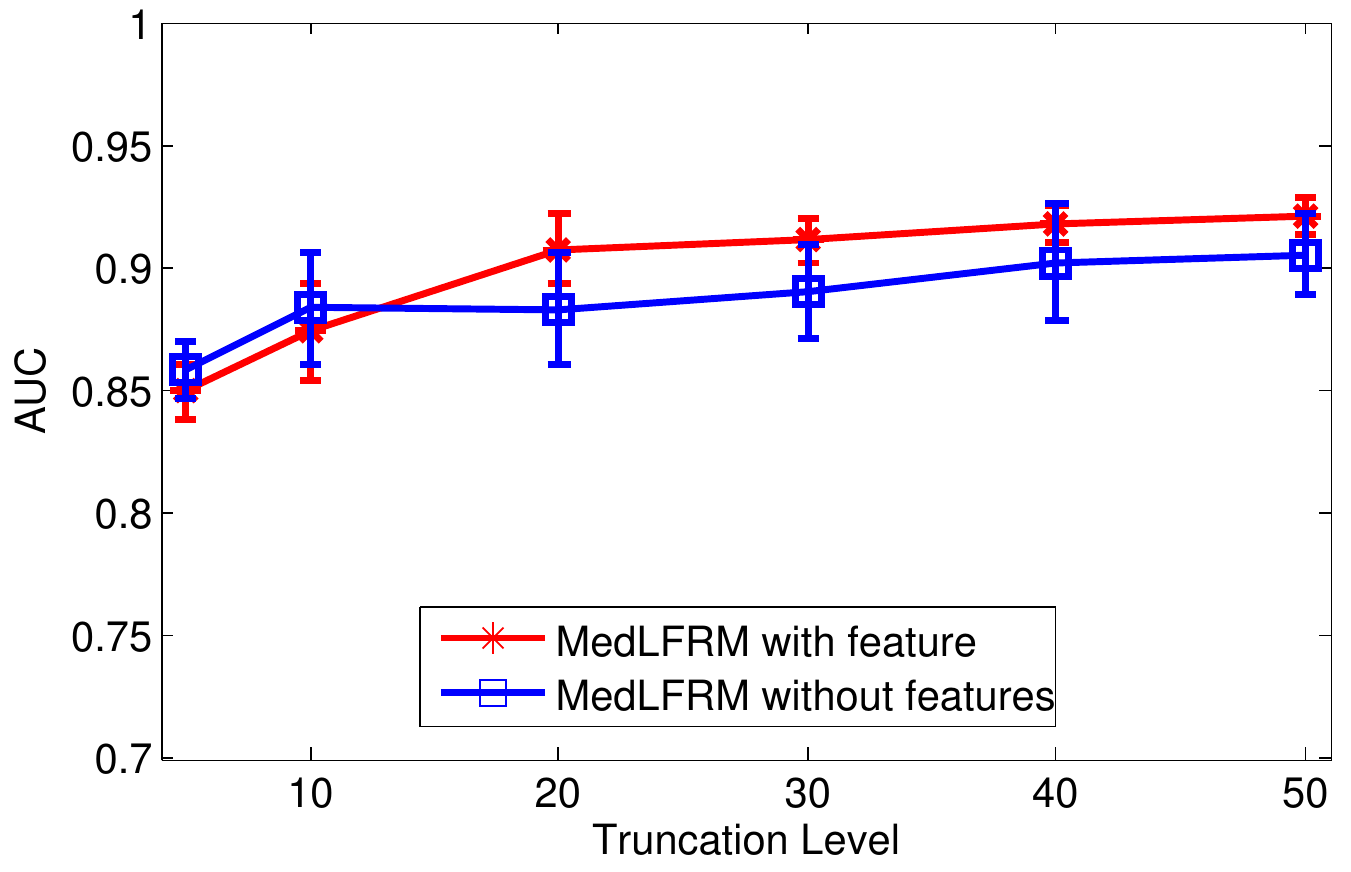}\vspace{-.3cm}
\caption{AUC scores of MedLFRM with and without input features on the countries dataset.}
\label{fig:NationComparison}\vspace{-.4cm}
\end{wrapfigure}

Fig.~\ref{fig:NationComparison} shows the performance of MedLFRM on the countries dataset when using and not using input features. We consider the global setting.
Here, we also study the effects of truncation level $K$. We can see that in general using input features can boost the performance.
Moreover, even if using latent features only, MedLFRM can still achieve very competitive performance, better than the performance of
the likelihood-based LFRM that uses both latent features and input features. Finally,
it is sufficient to get good performance by setting the truncation level $K$ to be larger than 40.
We set $K$ to be 50 in the experiments.

\begin{figure*}[t]\vspace{-.2cm}
\centering
{\hfill\subfigure[]{\includegraphics[height=1.2in, width=1.5in]{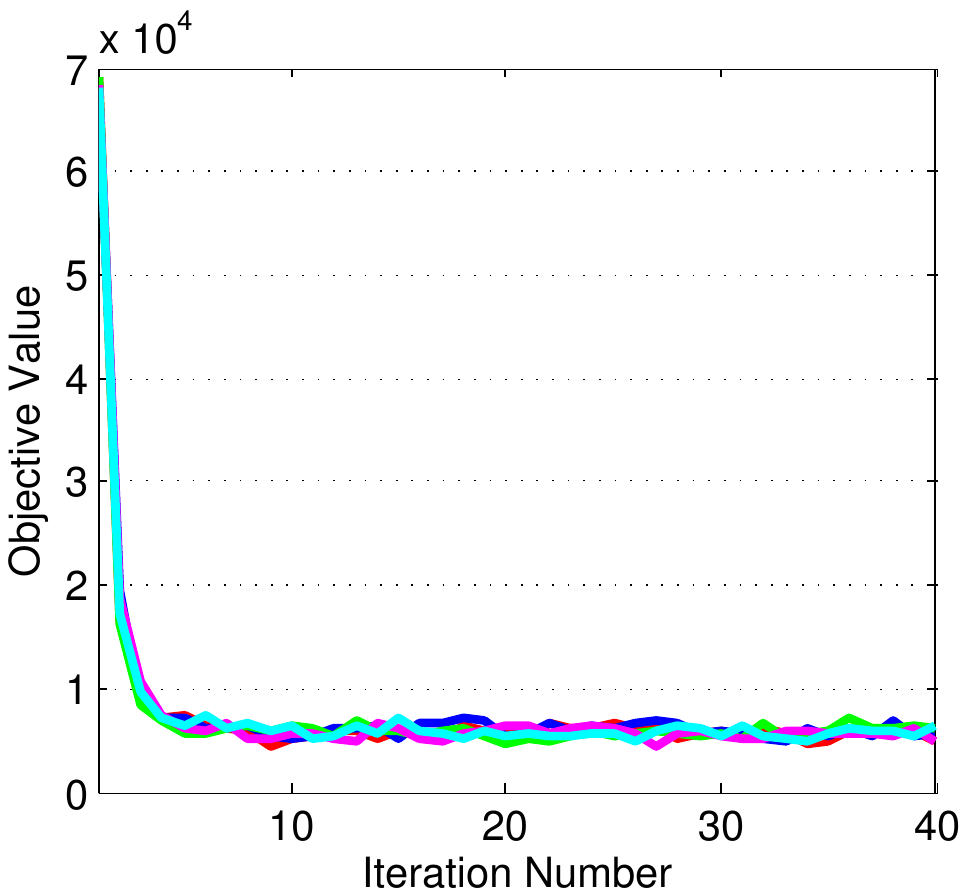}}\hfill
\subfigure[]{\includegraphics[height=1.14in, width=1.5in]{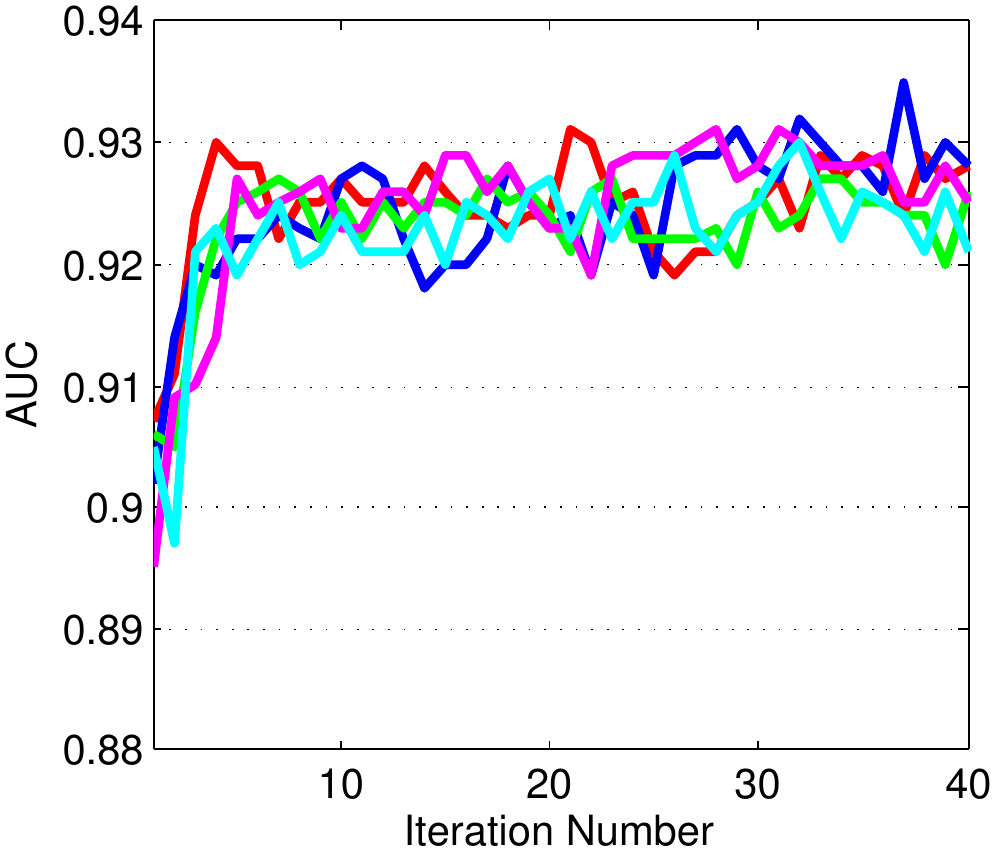}}\hfill
\subfigure[]{\includegraphics[height=1.2in, width=1.5in]{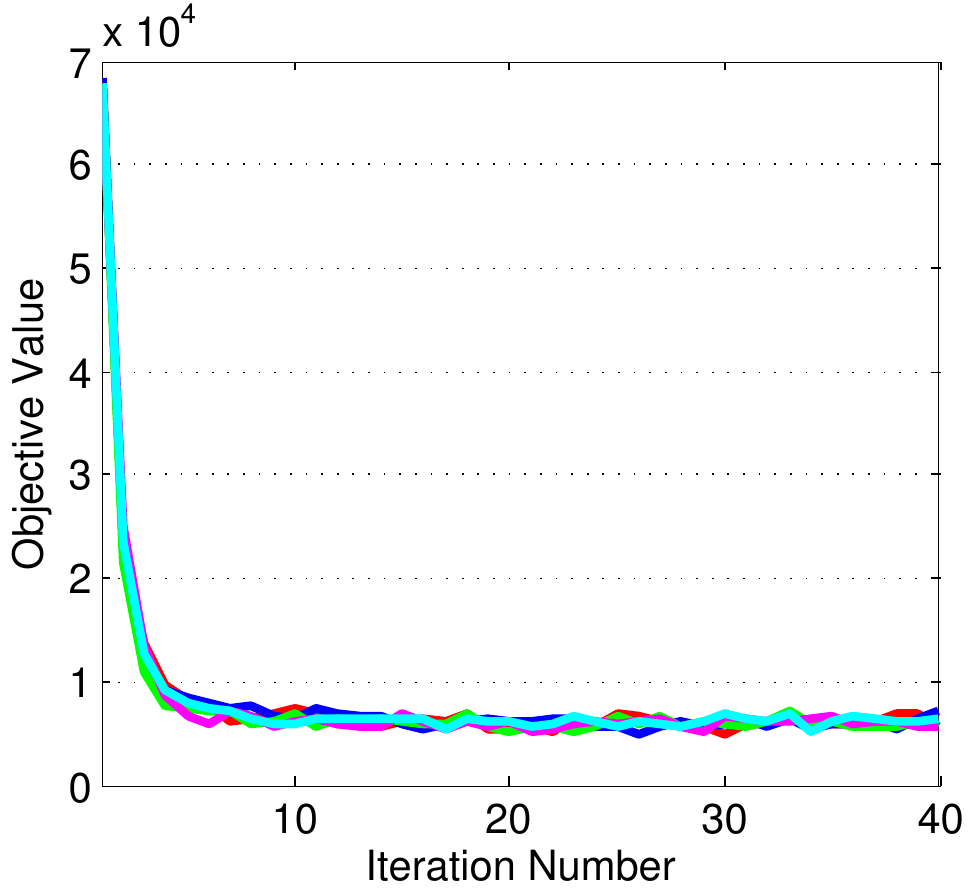}}\hfill
\subfigure[]{\includegraphics[height=1.14in, width=1.5in]{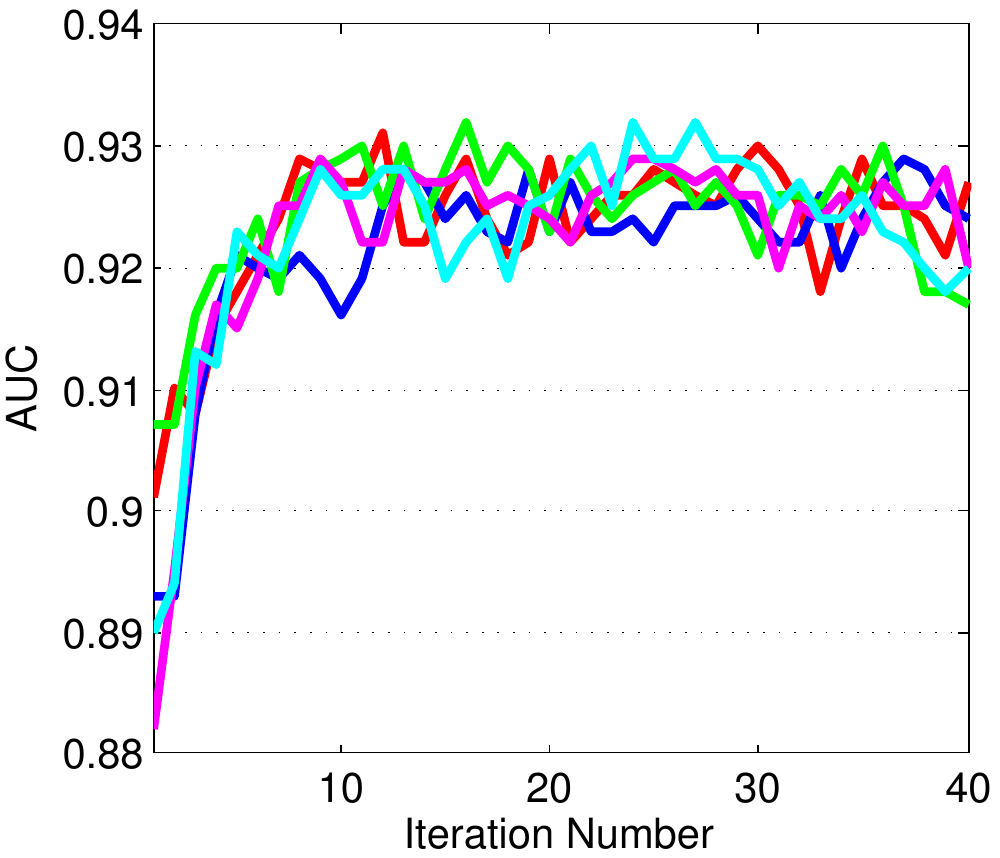}}\hfill} \vspace{-.2cm}
\caption{(a-b) Objective values and test AUC during iterations for MedLRFM; and (c-d) objective values and test AUC during iterations for Bayesian MedLRFM on the countries dataset with 5 randomly initialized runs.}
\label{fig:GNationStatble}\vspace{-.3cm}
\end{figure*}

\vspace{-.1cm}
\subsubsection{Predicting NIPS coauthorship}

The second experiments are done on a NIPS coauthorship dataset which contains a list of papers and authors from NIPS 1-17.\footnote{The empirical probability of forming a link is about $0.02$, again imbalanced.
We tried the same strategy as for Kinship by using different $C$ values, but did not observe obvious difference from that by using a common $C$.
$K=80$ is sufficient for this network.}
To compare with LFRM~\cite{Miller:nips09}, we use the same dataset which
contains 234 authors who had published with the most other people.
To better fit the symmetric coauthor link data, we restrict our models to be symmetric as in~\cite{Miller:nips09}, i.e., the posterior mean of $W$ is a symmetric matrix.
For MedLFRM and BayesMedLFRM, this symmetry constraint can be easily satisfied
when solving the SVM problems~(\ref{eq:PrimalSVMNormal}) and (\ref{eq:PrimalSVMGammaNormal}). To see the effects of the symmetry constraint, we also report the results of
the asymmetric MedLFRM and asymmetric BayesMedLFRM, which do not impose the symmetry constraint on the posterior mean of $W$.
As in~\cite{Miller:nips09}, we train the model on $80\%$ of the data and use the remaining data for test.

\begin{table}[t] \vspace{-.1cm}
\caption{AUC on the NIPS coauthorship data.}
\label{table:nips}\vspace{-.5cm}
\begin{center}
{\begin{tabular}{|c|c|}
\hline
MMSB        & 0.8705 $\pm$ 0.0130  \\
IRM		    & 0.8906 $\pm$ ---  \\
LFRM rand   & 0.9466 $\pm$ ---  \\
LFRM w/ IRM & 0.9509 $\pm$ --- \\
\hline
MedLFRM                  & {\bf 0.9642} $\pm$ 0.0026 \\
BayesMedLFRM             & {\bf 0.9636} $\pm$ 0.0036 \\
Asymmetric MedLFRM   	 & 0.9140 $\pm$ 0.0130 \\
Asymmetric BayesMedLFRM	 & 0.9146 $\pm$ 0.0047 \\
\hline
\end{tabular}} \vspace{-.6cm}
\end{center}
\end{table}

Table~\ref{table:nips} shows the results, where the results of LFRM, IRM and MMSB were reported in~\cite{Miller:nips09}. Again, we can see that using the discriminative max-margin training,
the symmetric MedLFRM and BayesMedLFRM outperform all other likelihood-based methods, using either latent feature or latent class models;
and the full Bayesian MedLFRM model performs comparably with MedLFRM while avoiding tuning the hyper-parameter $C$.
Finally, the asymmetric MedLFRM and BayesMedLFRM models perform much worse than their symmetric counterpart models,
but still better than the latent class models.

\vspace{-.1cm}
\subsubsection{Stability and Running Time}

Fig.~\ref{fig:GNationStatble} shows the change of training objective function as well as the test AUC scores
on the countries dataset during the iterations for both MedLFRM and BayesMedLFRM.
For MedLFRM, we report the results with the best $C$ selected via cross-validation. We can see that the
variational inference algorithms for both models converge quickly to a particular region. Since we use sub-gradient
descent to update the distribution of $Z$ and the subproblems of solving for $p(\Theta)$ can in practice only be approximately solved,
the objective function has some disturbance, but within a relatively very small interval.
For the AUC scores, we have similar observations, namely, within several iterations, we could have very good link prediction performance.
The disturbance is again maintained within a small region, which is reasonable for our approximate inference algorithms. Comparing the two models,
we can see that BayesMedLFRM has similar behaviors as MedLFRM, which demonstrates the effectiveness of using full-Bayesian techniques
to automatically learn the hyper-parameter $C$.
We refer the readers to~\cite{Zhu:icml12} for more results on the kinship dataset,
from which we have the same observations. We omit the results on the NIPS dataset for saving space.

\begin{figure}\vspace{-.2cm}
\centering
{\hfill
\subfigure[]{\includegraphics[width=0.24\textwidth]{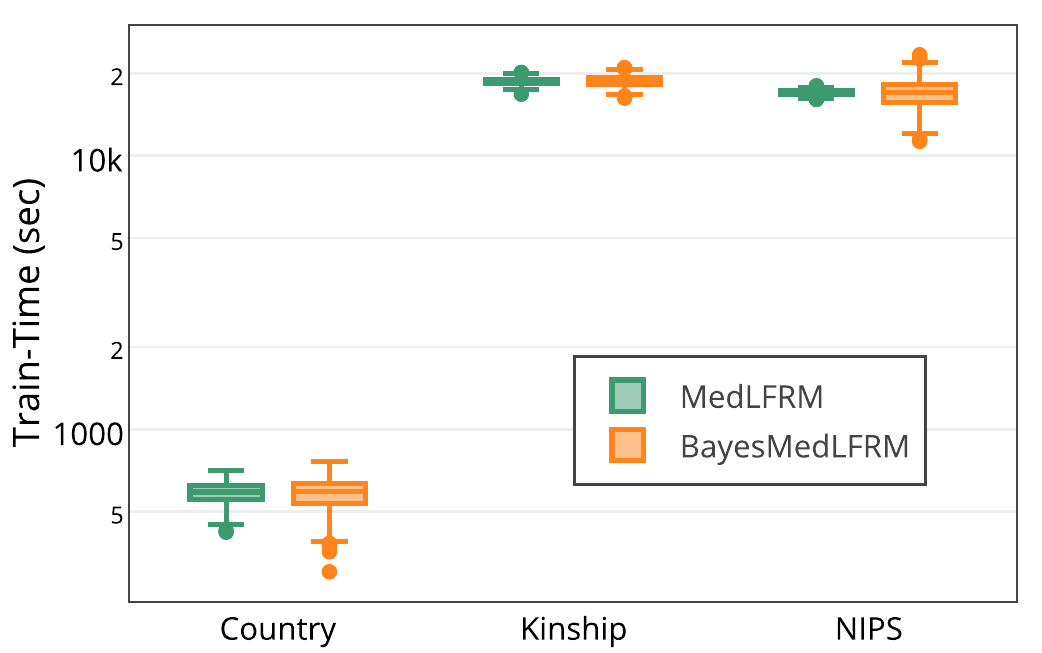}}
\hfill
\subfigure[]{\includegraphics[width=0.24\textwidth]{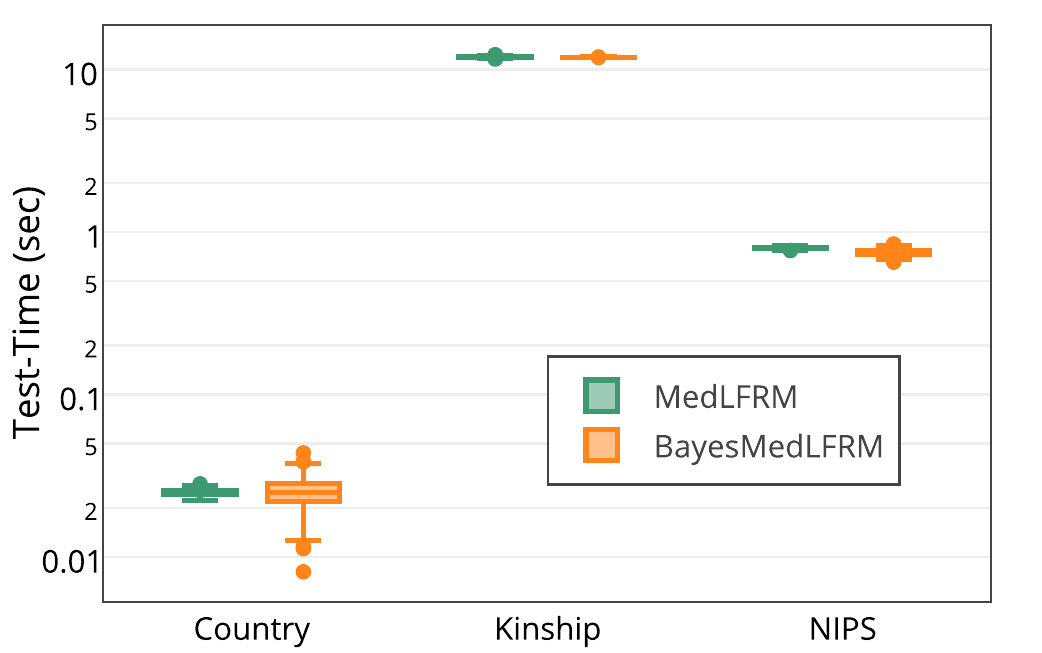}}
\hfill}\vspace{-.3cm}
\caption{Training and test time on different datasets.} 
\label{fig:Time}\vspace{-.3cm}
\end{figure}

Fig.~\ref{fig:Time} shows the training time and test time of MedLFRM and BayesMedLFRM\footnote{We do not compare with competitors whose implementation is not available.}
on all the three datasets. For MedLFRM, we show the single run with the optimal parameter $C$,
selected via inner cross-validation. We can see that using Bayesian inference,
the running time does not increase much, being generally comparable with that of MedLFRM.
But since MedLFRM needs to select the hyper-parameter $C$, it will need much more time than BayesMedLFRM
to finish the entire training on a single dataset. Table~\ref{table:time-svm-vi} further compares the time on learning SVMs (i.e., $p(\Theta)$) and the time on variational inference of $p(\nuv, Z)$. We can see that the time consumed in solving for $p(\nuv, Z)$ is the bottleneck for acceleration.

\iftrue
\begin{table}
\caption{Average Training Time for $p(\Theta)$ and $p(\nuv, Z)$}\vspace{-.3cm}
\centering
\begin{tabular}{|c|c|c|c|}
\hline
  & Countries & Kinship & NIPS  \\\hline
$p(\Theta)$     & 108.34    & 3978    & 1343\\
$p(\nuv, Z)$      & 480.73    & 14735   & 15699\\\hline
\end{tabular}
\label{table:time-svm-vi}\vspace{-.4cm}
\end{table}
\fi

\vspace{-.1cm}
\subsubsection{Sensitivity Analysis}

We analyze the sensitivity of MedLFRM with respect to the regularization parameter $C$, using the NIPS dataset as an example. 

\begin{figure}[h]\vspace{-.2cm}
\centering
\includegraphics[height=1.6in, width=0.78\linewidth]{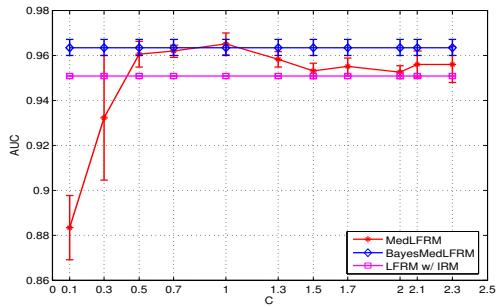}\vspace{-.3cm}
\caption{AUC of MedLFRM on NIPS dataset when the parameter $C$ changes, comparing with BayesMedLFRM and LFRM w/ IRM which are not affected by $C$.}
\label{fig:sensitivity_c_nips}\vspace{-.1cm}
\end{figure}

Fig.~\ref{fig:sensitivity_c_nips} shows the performance of MedLFRM when the hyper-parameter $C$ changes from $0.1$ to $2.3$, with comparison to BayesMedLFRM and LFRM (See Table 2 for exact AUC scores) --- BayesMedLFRM automatically infers $C$ while LFRM does not have a similar hyper-parameter. We can see that $C$ is an important parameter that affects the performance of MedLFRM. In the experiments, we used cross-validation to select $C$, which needs to learn and compare multiple candidate models.
In contrast, BayesMedLFRM avoids tuning $C$ by automatically inferring the posterior mean and variance of $\Theta$ (the effective $C$ equals to $\frac{1}{\ep[\tau]}$, where $\ep[\tau]$ is updated using Eq.~(\ref{eq:postmean-NG})). BayesMedLFRM does not need to run for multiple times. As shown in Fig.~\ref{fig:Time}, the running time of BayesMedLFRM is comparable to the single run of MedLFRM. Thus, BayesMedLFRM can be more efficient in total running time. In all the experiments, we fixed the Normal-Gamma prior to be a weakly informative prior (See Section~\ref{sec:multirelation}), which still leads to very competitive prediction performance for BayesMedLFRM. 

\subsubsection{Sparsity}

\begin{wraptable}{r}{.26\textwidth} \vspace{-1.8cm}
\caption{Sparsity of $\psi$.}\vspace{-.3cm}
\centering
\begin{tabular}{|c|c c|}
\hline
$K$  & number  & ratio (\%)   \\\hline
80     & 1071  & 0.058 \\
120      & 1514   & 0.055 \\
160 & 1765 & 0.048 \\ \hline
\end{tabular}
\label{table:sparsity}\vspace{-.3cm}
\end{wraptable}

We analyze the sparsity of the latent features. For our variational methods, the posterior mean of $\Zv$ (i.e., $\psi$) is not likely to have zero entries. Here, we define the sparsity as ``less-likely to appear" --- if the posterior probability of a feature is less than 0.5, we treat it as less-likely to appear. Table~\ref{table:sparsity} shows the number of ``non-zero" entries of $\psi$ and the ratio when the truncation level $K$ takes different values on the NIPS dataset. We can see that only very few entries have a higher probability to be active (i.e., taking value 1) than being inactive. Furthermore, the sublinear increase against $K$ suggests convergence. Finally, we also observed that the number of active columns (i.e., features) converge when $K$ goes larger than 120. For example, when $K=160$, about 134 features are active in the above sense.

\vspace{-.1cm}
\subsection{Results with Stochastic Algorithms}
\vspace{-.1cm}

\begin{table}\vspace{-.15cm}
\caption{Running Time and AUC of the stochastic MedLFRM on Kinship (Single) and NIPS datasets}
\label{table:svi-time}\vspace{-.2cm}
\centering
\begin{tabular}{|c|cc|}
\hline
                     & Kinship             & NIPS\\\hline
Average Running Time & 115.34              & 2425.33 \\
AUC			         & 0.9543 $\pm$ 0.0088 & 0.9596 $\pm$ 0.0075  \\
Average Speed-up     & 6.24x               & 6.15x \\\hline

\end{tabular}\vspace{-.2cm}
\end{table}

We now demonstrate the effectiveness of our stochastic algorithms on dealing with large-scale networks, which are out of reach for the batch algorithms.

\vspace{-.15cm}
\subsubsection{Results on Small Datasets}

We first analyze how well the stochastic methods perform by comparing with the batch algorithms on the Kinship and NIPS datasets.

For the stochastic methods, we randomly select a fixed size $N^\prime$ of entities, and for each entity we sample $M^\prime$ associated links\footnote{In the case where $M^\prime$ is larger than the number of associated links of entity $i$, we use the original update algorithm for $p(Z_i)$.} at each iteration. For the Kinship dataset, we set $N^\prime = 10$ and $M^\prime = 50$. For the NIPS dataset, $N^\prime = 50$ and $M^\prime = 50$. A reasonable sub-network size is selected for stability and convergence to the best AUC. We use $\kappa_\gamma = 0$, and $\kappa_\psi = 0.5$ in both settings.
Table~\ref{table:svi-time} shows the results. We can see that our stochastic algorithms have a significant speed-up while maintaining the good performance of the original model. This speed up increases when we choose smaller sub-networks compared with the total network, allowing us to make inference on larger networks in a reasonable time.

\begin{table}\vspace{-.2cm}
\caption{Network Properties of AstroPh and CondMat}\vspace{-.3cm}
\centering
\begin{tabular}{|c|c|c|c|c|}
\hline
Name & \# nodes & \# links & Max degree & Min degree \\\hline
AstroPh & 17,903 & 391,462 & 1,008 & 2\\
CondMat & 21,363 & 182,684 & 560 & 2\\\hline
\end{tabular}
\label{table:prop-ap-cm}
\end{table}

\begin{figure}[t]
\centering
\includegraphics[width=0.7\linewidth]{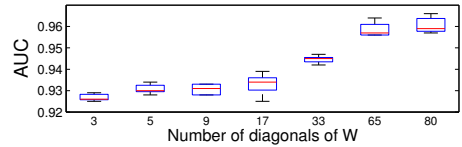}\vspace{-.2cm}
\caption{AUC of (symmetric) MedLFRM on NIPS dataset with $W$ having various numbers of non-zero diagonals.}
\label{fig:nipsw}\vspace{-.4cm}
\end{figure}

\begin{figure}[t]\vspace{-.4cm}
\centering
\includegraphics[width=0.88\linewidth]{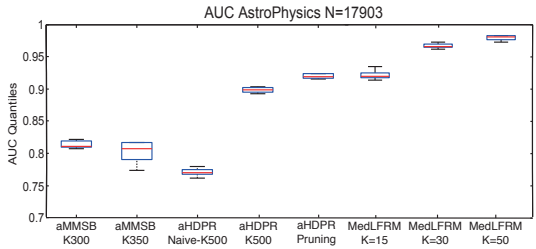}\vspace{-.2cm}
\caption{AUC results on the AstroPh dataset. The results of baseline methods are cited from \cite{Kim:2013}.}
\label{fig:ap-comp}\vspace{-.1cm}
\end{figure}

\begin{table}[t]
\caption{AUC and Training Time on AstroPh and CondMat}
\label{table:auc-ap-cm}\vspace{-.3cm}
\centering
\begin{tabular}{|c|c|c|c|}
\hline
Dataset & K & Test AUC & Training Time(s) \\\hline
\multirow{3}{*}{AstroPh} & 15 & $0.9258 \pm 0.0010$ & $1094 \pm 207$ \\
\cline{2-4}
						 & 30 & $ 0.9648 \pm 0.0004$ & $5853 \pm 382$\\
\cline{2-4}
						 & 50 & $0.9808 \pm 0.0004$ & $19954 \pm 224$\\\hline
\multirow{3}{*}{CondMat} & 30 & $0.8912 \pm 0.0027$ & $2751 \pm 321$\\
\cline{2-4}
						 & 50 & $0.9088 \pm 0.0075$ & $10379 \pm 283$\\
\cline{2-4}
						 & 70 & $0.9212 \pm 0.0027$ & $27551 \pm 392$\\\hline
\end{tabular}\vspace{-.4cm}
\end{table}

\begin{figure}[t]
\centering
\includegraphics[width=0.88\linewidth]{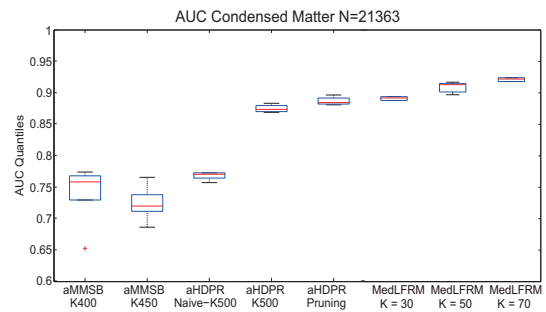}\vspace{-.6cm}
\caption{AUC results on the CondMat dataset. The results of baseline methods are cited from \cite{Kim:2013}.}
\label{fig:cm-comp}\vspace{-.3cm}
\end{figure}

\vspace{-.15cm}
\subsubsection{Results on Two Large Networks}

We then present the results on 
the Arxiv Astro Physics collaboration network~(AstroPh) and the Arxiv Condensed Matter collaboration network~(CondMat). We use the same settings as \cite{Kim:2013}, extracting the largest connected component, where AstroPh contains 17,903 nodes and CondMat contains 21,363 nodes. Table \ref{table:prop-ap-cm} describes the statistics of the two networks, where we count a collaboration relationship as two directed links. 
For AstroPh, we set $N^\prime = M^\prime = 500$, $\kappa_\gamma=0$, and $\kappa_\phi=0.2$; for CondMat, we set $N^\prime = 750$, $M^\prime$ be the maximum degree, $\kappa_\gamma = 0$, and $\kappa_\psi = 0.2$. 
To further increase inference speed, we restrict $W$ to contain only non-zero entries on the diagonal, superdiagonal and subdiagonal, which decreases the non-zero elements of $W$ from $O(K^2)$ to $O(K)$ and the complexity of computing $p(Z)$ to $O(|\mathcal{I}| K)$. Our empirical studies show that this restriction still provides good results\footnote{On the NIPS dataset, we can obtain an average AUC of 0.926 when we impose the diagonal plus off-diagonal restriction on $W$ when $K=80$. Fig.~\ref{fig:nipsw} shows more results when we
gradually increase the number of off-diagonals under 
the same setting as in Table \ref{table:svi-time} with $K$ fixed at $80$. Note that increasing $K$ could possibly increase the AUC for each single setting.}. 
Since the two networks are very sparse, we randomly select 90\% of the collaboration relationships as positive examples and non-collaboration relationships as negative examples for training, such that the number of negative examples is almost 10 times the number of positive examples. Our test set contains the remaining 10\% of the positive examples and the same number of negative examples, which we uniformly sample from the negative examples outside the training set. This test setting is the same as that in~\cite{Kim:2013}. 

Table \ref{table:auc-ap-cm} shows the AUC and training time on both datasets. Although the sub-networks we choose at each iteration are different during each run, the AUC and training time are stable with a small deviation. We choose different values for $K$, which controls the number of latent features for each entity. Under a fixed number of iterations, smaller $K$'s allow us to make a reasonable inference faster, while larger $K$'s give better AUC scores. We compare with the state-of-the-art nonparametric latent variable models, including assortative MMSB (aMMSB)~\cite{gopalan2013efficient} and assortative HDP Relational model~(aHDPR)~\cite{Kim:2013}, a nonparametric generalization of aMMSB.
Fig.~\ref{fig:ap-comp} and Fig.~\ref{fig:cm-comp} present the test AUC scores, where the results of aMMSB and aHDPR are cited from \cite{Kim:2013}.
We can see that our MedLFRM with smaller $K$'s have comparable AUC results to that of the best baseline (i.e., aHDPR with pruning), while we achieve significantly better performance when $K$ is relatively large (e.g., $50$ on both datasets or $70$ on the CondMat dataset).

\subsubsection{Sensitivity Analysis for $N^\prime$ and $M^\prime$}

We analyze the sensitivity of the stochastic algorithm of MedLFRM with respect to the network size, namely the number of entities sampled per sub-network~($N^\prime$) and the number of links sampled from each entity ($M^\prime$), using the AstroPh dataset as an example.

\begin{figure*}[t]\vspace{-.2cm}
\centering
{\hfill\subfigure[]{\includegraphics[height=1in, width=2in]{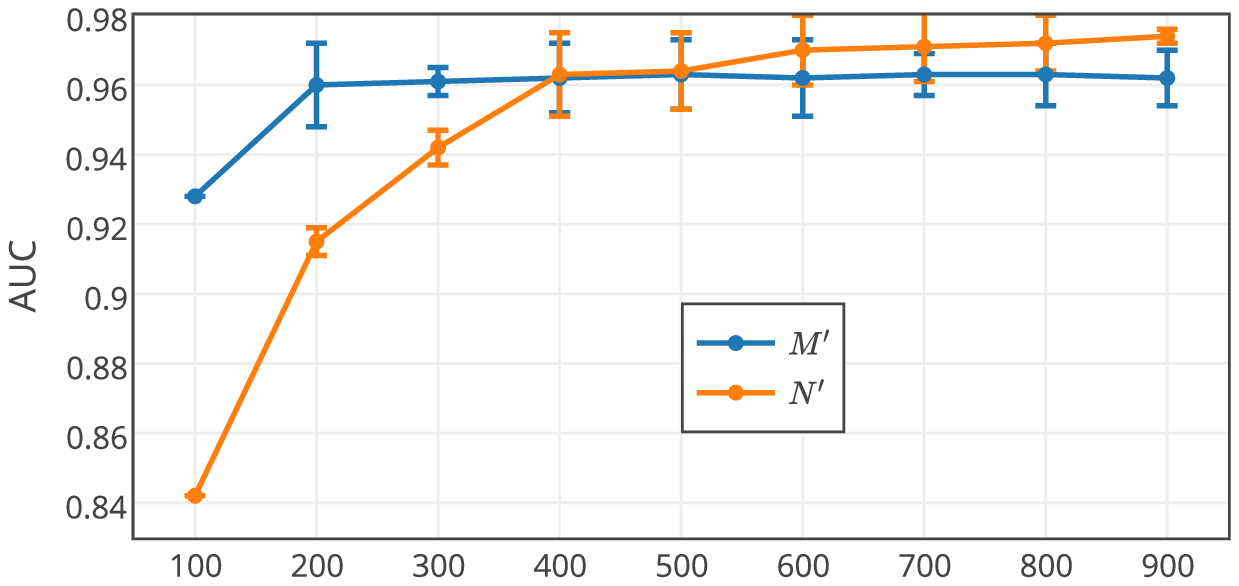}}\hfill
\subfigure[]{\includegraphics[height=1in, width=2in]{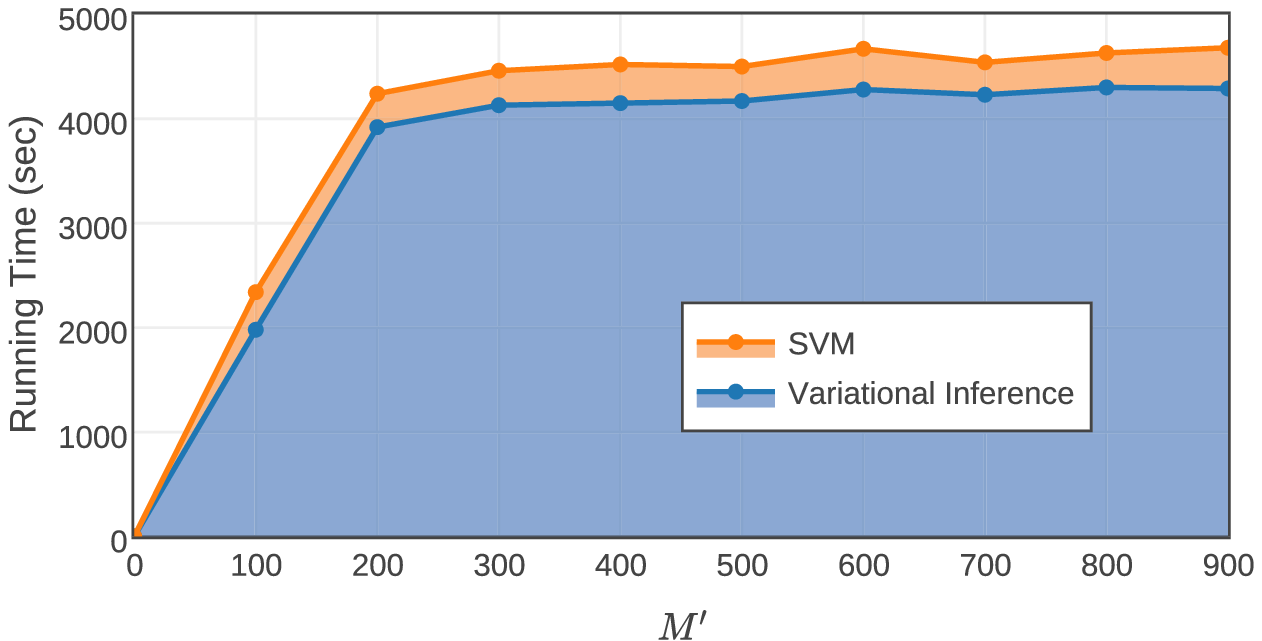}}\hfill
\subfigure[]{\includegraphics[height=1in, width=2in]{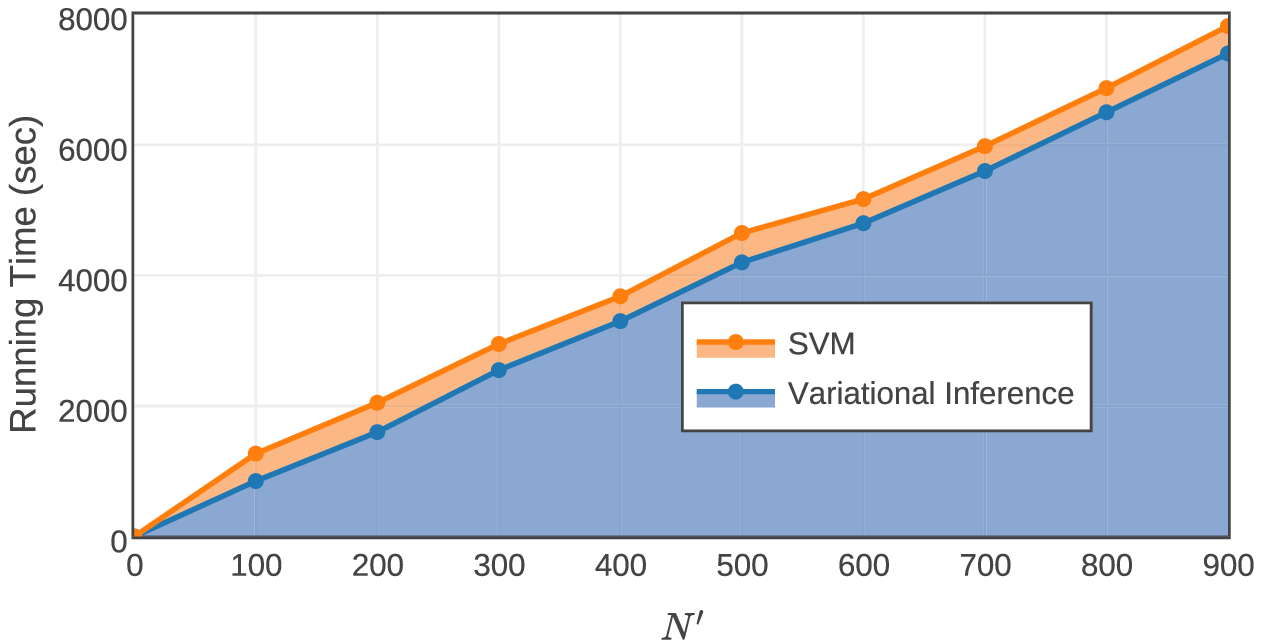}}\hfill
} \vspace{-.3cm}
\caption{On the AstroPh dataset where (a) Test AUC values with \emph{(a.i)} $N^\prime=500$ and $M^\prime$ from $100$ to $900$, and \emph{(a.ii)} $M^\prime=500$ and $N^\prime$ from $100$ to $900$; (b) Training time with \emph{(a.i)}; (c) Training time with \emph{(a.ii)}.}
\label{fig:ap-sensitivity}\vspace{-.4cm}
\end{figure*}

Fig.~\ref{fig:ap-sensitivity} shows the AUC and training time of the stochastic MedLFRM in two scenarios: ({\bf 1}) $N^\prime$ changes from $100$ to $900$, while $M^\prime = 500$; and ({\bf 2}) $N^\prime = 500$, while $M^\prime$ changes from $100$ to $900$. Larger $N^\prime$s and $M^\prime$s indicate a larger sub-network sampled at each iteration. We also report the training time of SVM (i.e., computing $p(\Theta)$) and variational inference (i.e., computing $p(\nuv, Z)$) respectively. We fix all other parameters in these settings.

We can see that the training time increases as we sample more links in a sub-network. The training time consumed in the SVM-substep is smaller than that of inferring $p(\nuv, Z)$ similar as observed in Table~\ref{table:time-svm-vi}.
The training time consumed in variational inference is almost linear to $N^\prime$, which is reasonable since the expected number of links sampled scales linearly with $N^\prime$; this property, however, is not observed when we change $M^\prime$. This is mainly because only $11\%$ of the entities are associated with more than 200 training links, and only the links associated with these entities are affected by larger $M^\prime$s. Thus, increasing $M^\prime$ in this case does not significantly increase the number of sampled links as well as the training time.

We can also see a low AUC when $N^\prime$ or $M^\prime$ is 100, where the sub-network sampled at each iteration is not large enough to rule out the noise in relatively few iterations, therefore leading to poor AUC results. In contrast, if the sampled sub-networks exceed a certain size (e.g., $N^\prime > 500$ or $M^\prime > 500$), the AUC will not have significant increase. Therefore, there exists a trade-off between training time and accuracy, determined by the sizes of the sampled sub-networks, and choosing the most suitable network size would require some tuning in practice.

\subsubsection{Results on a Massive Dataset}

Finally, we demonstrate the capability of our models on the US Patent dataset, a massive citation network containing 3,774,768 patents and a total of 16,522,438 citations. For 1,803,511 of the patents, we have no information about their citations, although they are cited by other patents.
We construct our training and test set as follows: we include all the edges with observed citations, and uniformly sample the remaining edges without citations. We extract 21,796,734 links (which contain all the positive links, while the negative links are randomly sampled), and uniformly sample 17,437,387 links as training links. We set $N^\prime=50,000$ and $M^\prime$ equals to the maximum degree, that is, all the links associated with an entity are selected.

For baseline methods, we are not aware of any sophisticated models that have been tested on this massive dataset for link prediction. Here, we present a first try and compare with the proximity-measure based methods~\cite{liben_nowell}, including common neighbors (CN), Jaccard cofficient, and Katz.
Since the citation network is directed, we consider various possible definitions of common neighbors~\footnote{Let $\mathcal{C}_{i0} \triangleq \{k: (i, k) \in \mathcal{I} \}$ be the set of out-links of entity $i$ and $\mathcal{C}_{i1} \triangleq \{k: (k, i) \in \mathcal{I} \}$ be the set of in-links.
Then, common neighbors can be defined as: 1) $\mathcal{C}_{i0} \cap \mathcal{C}_{j0}$: children; 2) $\mathcal{C}_{i1} \cap \mathcal{C}_{j1}$: parents;
3) $\mathcal{C}_{i0} \cap \mathcal{C}_{j1}$ (or $\mathcal{C}_{i1} \cap \mathcal{C}_{j0}$): intermediate nodes on the length-2 paths from $i$ to $j$ (or from $j$ to $i$);
or 4) union of the above sets.} 
as well as Jaccard coefficient and report their best AUC scores.
The Katz measure~\cite{katz1953new} is defined as the summation over the collection of paths from entity $i$ to $j$, exponentially damped by path length. We set the damped coefficient to be $0.5$, which leads to the best AUC score.

From Table \ref{table:patent}, we can observe that our latent feature model achieves a significant improvement on AUC scores over the baseline methods with a reasonable running time. Though the simple methods, such as CN and Jaccard, are very efficient, our method achieves significantly better AUC than the Katz method with less running time (e.g., a half when $K=30$). We can achieve even better results (e.g., $K=50$) in about 10 hours on a standard computer. The gap between training and testing AUC can partially be explained by the nature of the data---the information concerning the citation of nearly $50\%$ of the patents is missing; when we sample the negative examples, we are making the oversimplified assumption that these patents have no citations; it is likely that the training data generated under this assumption is deviated from the ground truth, hence leading to a biased estimate of the citation relations.

\section{Conclusions and Discussions}
We present a discriminative max-margin latent feature relational model for link prediction.
Under a Bayesian-style max-margin formulation, our work naturally integrates the ideas of Bayesian nonparametrics
which can infer the unknown dimensionality of a latent social space.
Furthermore, we present a full Bayesian formulation, which avoids tuning regularization constants.
For posterior inference and learning, we developed efficient stochastic variational methods, which can scale up to real networks with millions of entities. Our empirical results on a wide range of real networks demonstrate the
benefits inherited from both max-margin learning and Bayesian methods.

Our current analysis is focusing on static network snapshots. For future work, we are interested in learning more flexible latent feature relational models to deal with dynamic networks and reveal more subtle network evolution patterns. Moreover, our algorithms need to specify a truncation level. Though a sufficiently large truncation level guarantees to infer the optimal latent dimension, it may waste computation cost. The truncation-free ideas~\cite{Wang:2012} will be valuable to explore to dynamically adjust the latent dimension.

\begin{table}[t]
\caption{Results on the US Patent dataset.}\vspace{-.2cm}
\centering
\setlength{\tabcolsep}{2.5pt}
\begin{tabular}{|c|c|c|c|c|}
\hline
Method & K & Test AUC & Train AUC & Running Time (s) \\\hline
\multirow{3}{*}{MedLFRM} & 15 & $0.653 \pm 0.0033$ & $0.796 \pm 0.0057$ & $2787 \pm 132$ \\\cline{2-5}
						 & 30 & $0.670 \pm 0.0029$ &        $0.831 \pm 0.0042$        & $10342 \pm 648$\\\cline{2-5}
						 & 50 & $0.685 \pm 0.0035$ &        $0.858 \pm 0.0076$            & $37860 \pm 1224$\\\hline
CN                         & -  & $0.619$            &       ---            & $87.00 \pm 2.58$ \\\hline

Jaccard                     & -  & $0.618$            &       ---            & $52.15 \pm 2.46$ \\\hline

Katz & - & $0.639$ & --- & $ 21975  \pm 259 $ \\\hline
\end{tabular}
\label{table:patent}\vspace{-.4cm}
\end{table}

\vspace{-.15cm}
\ifCLASSOPTIONcompsoc
  \section*{Acknowledgments}
\else
  \section*{Acknowledgment}
\fi
\vspace{-.1cm}

{This work is supported by the National 973 Basic Research Program of China (Nos. 2013CB329403, 2012CB316301), National NSF of China (Nos. 61322308, 61332007), Tsinghua Initiative Scientific Research Program (No. 20141080934)}.


%
\vspace{-.2cm}
\bibliographystyle{plain}
\bibliography{dpmed}

\appendices
\vspace{-.2cm}
\section*{Appendix A. Evaluating KL-divergence}
By the mean-field assumption, we have the form\vspace{-.2cm}
\begin{eqnarray}
\KL(p(\nuv, Z) &\Vert& p_0(\nuv, Z)) = \KL(p(\nuv | \gammav) \Vert p_0(\nuv)) \nonumber \\
&& + \sum_{i=1}^N \ep_{p(\nuv)}\left[ \KL(p(Z_i | \psiv_i) \Vert p_0(Z_i|\nuv)) \right], \nonumber \vspace{-.35cm}
\end{eqnarray}
with each term evaluated as
$\KL(p(\nuv|\gammav) \Vert p_0(\nuv)) = \sum_{k=1}^K \big( (\gamma_{k1}-\alpha) \ep_p[ \log \nu_k ] + (\gamma_{k2}-1) \ep_p[ \log (1 - \nu_k) ] - \log B(\gamma_k) \big) - K \log \alpha$ and
$\ep_{p}\left[ \KL(p(Z_i | \psiv_i) \Vert p_0(Z_i|\nuv)) \right] = \sum_{k=1}^K (  - \psi_{ik} \sum_{j=1}^k \ep_p[\log \nu_j]
 - (1-\psi_{ik})\ep_p [ \log(1 - \prod_{j=1}^k \nu_j) ] - \mathcal{H}(p(Z_{ik} | \psi_{ik}))  )$,
where $\ep_p[ \log v_j ] = \psi(\gamma_{j1}) - \psi(\gamma_{j1} + \gamma_{j2}), ~ \ep_p[ \log (1-v_j) ] = \psi(\gamma_{j2}) - \psi(\gamma_{j1} + \gamma_{j2})$, $\psi(\cdot)$ is the digamma function, $\mathcal{H}(p(Z_{ik}|\psi_{ik}))$ is the entropy of the Bernoulli distribution $p(Z_{ik}|\psi_{ik})$, and $B(\gamma_k) = \frac{\Gamma(\gamma_{k1})\Gamma(\gamma_{k2})}{\Gamma(\gamma_{k1}+\gamma_{k2})}$.
All the above terms can be easily computed, except the term $\ep_p[ \log(1 - \prod_{j=1}^k \nu_j) ]$. Here, we adopt the multivariate lower bound~\cite{YWTeh:aistats09}:\vspace{-.2cm}
\setlength\arraycolsep{1pt} \begin{eqnarray}
&& \ep_p[ \log(1 -  \prod_{j=1}^k \nu_j) ] \geq \mathcal{H}(q_{k.}) + \sum_{m=1}^k q_{km} \psi(\gamma_{m2}) \nonumber \\
&& ~~~~~~~~~~~ + \sum_{m=1}^{k-1} \zeta_1 \psi(\gamma_{m1})
 - \sum_{m=1}^k \zeta_2 \psi(\gamma_{m1}+\gamma_{m2}), \nonumber \vspace{-.35cm}
\end{eqnarray}
where the variational parameters $q_{k.}=(q_{k1} \cdots q_{kk})^\top$ belong to the $k$-simplex, $\mathcal{H}(q_{k.})$ is the entropy of $q_{k.}$, $\zeta_1 = \sum_{n=m+1}^k \! q_{kn}$ and $\zeta_2 = \sum_{n=m}^k  q_{kn}$.
The tightest lower bound is achieved by setting $q_{k.}$ to be the optimum value
$q_{km} \propto \exp( \psi(\gamma_{m2}) + \sum_{n=1}^{m-1} \psi(\gamma_{n1}) - \sum_{n=1}^m \psi(\gamma_{n1}+\gamma_{n2}))$. 
We denote the tightest lower bound by $\mathcal{L}_k^\nu$. Replacing the term $\ep_p[ \log(1 -  \prod_{j=1}^k \nu_j) ]$ with its lower bound $\mathcal{L}_k^\nu$,
we can have an upper bound of $\KL(p(\nuv, Z)\Vert p_0(\nuv, Z))$.

\vspace{-.15cm}
\section*{Appendix B: Variational Inference for Normal-Gamma Bayesian Model}

The variational inference is to find a distribution $p(\mu, \tau, \Theta)$ that solves problem~(\ref{eq:VarNormalGamma}).
We make the mean field assumption that $p(\mu, \tau, \Theta) = p(\mu, \tau) p(\Theta)$.
Then, we can get the update equation:
$p(W_{kk^\prime}) = \mathcal{N}(\Lambda_{kk^\prime}, \lambda^{-1} ),~p(\eta_d) = \mathcal{N}(\kappa_d, \lambda^{-1})$,
where $\Lambda_{kk^\prime} = \ep[\mu] + \lambda^{-1} \sum_{(i,j) \in \mathcal{I}} \omega_{ij}Y_{ij} \ep[Z_{ik}Z_{jk^\prime}],$
$\kappa_d = \ep[\mu] + \lambda^{-1} \sum_{(i,j) \in \mathcal{I}} \omega_{ij}Y_{ij} X_{ij}^d$, and $\lambda = \ep [\tau]$. Similar as in MedLFRM, the posterior mean can be obtained by solving a binary SVM subproblem~(\ref{eq:PrimalSVMGammaNormal}).

Then, minimizing the objective over $p(\mu, \tau)$ leads to the mean-field update equation $p(\mu, \tau) \propto p_0(\mu, \tau|\mu_0, n_0, \nu_0, S_0) \exp\left( - \Delta \right) $,
where $\Delta \triangleq - \ep[\log p_0(\Theta | \mu, \tau) ] =  \frac{\tau \ep \big[ \Vert W - \mu E\Vert_2^2 + \Vert \eta - \mu \ev\Vert_2^2\big] }{2} - \frac{c \log \tau}{2} + c^\prime$, where $c=K^2+D$ and $c^\prime = \frac{K^2+D}{2} \log 2\pi$ are constants. Doing some algebra, we can get
$\Delta 
 =   \frac{\tau (\ep[ S_W  + S_\eta] + K^2 (\bar{\Lambda} - \mu)^2 + D (\bar{\kappa} - \mu)^2  + \frac{c}{\lambda})}{2}  - \frac{c \log \tau}{2} + c^\prime$, where
$S_W = \sum_{kk^\prime} (W_{kk^\prime} - \bar{W})^2$ and~$S_\eta = \sum_d (\eta_d - \bar{\eta})^2$. Then, we can show that $p(\mu, \tau)$ is\vspace{-.2cm}
\begin{eqnarray}
p(\mu, \tau) = \mathcal{NG}( \tilde{\mu}, \tilde{n}, \tilde{\nu}, \tilde{S}), \nonumber \vspace{-.35cm}
\end{eqnarray}
where $\tilde{n} = n_0 + c$, $\tilde{\nu} = \nu_0 + c$, $\tilde{\mu} = \frac{K^2 \bar{\Lambda} + D \bar{\kappa} + n_0 \mu_0}{c + n_0}$, $\tilde{S} = \ep [S_{W} + S_{\eta}] + S_0 + \frac{n_0 (K^2 (\bar{\Lambda} - \mu)^2 + D (\bar{\kappa} - \mu)^2) }{c + n_0}.$
From $p(\mu, \tau)$, we can compute the expectation and variance as in Eq.~(\ref{eq:postmean-NG}), which are needed in updating $p(\Theta)$ and evaluating the objective function.

Now, we can evaluate the objective function. The KL-divergence in problem~(\ref{eq:VarNormalGamma}) is \vspace{-.2cm}
\begin{eqnarray}
\mathcal{L} 
 = && \frac{\tilde{\nu}}{2} \log \frac{\tilde{S}}{2} + \frac{\log \frac{\tilde{n}}{n_0}}{2} - \log \Gamma(\frac{\tilde{\nu}}{2}) - \frac{\nu_0}{2} \log \frac{S_0}{2} \nonumber \\
 && + \log \Gamma(\frac{\nu_0}{2}) + \frac{(\tilde{\nu} - \nu_0)\ep[ \log \tau ]}{2}  - \frac{(\tilde{S} - S_0)\ep[ \tau ]}{2} \nonumber \\
&& -  \frac{ \ep[ \tilde{n}\tau(\mu - \tilde{\mu})^2  - n_0\tau(\mu - \mu_0)^2 ]}{2}  \nonumber \\
&& + \frac{ K^2 (\log \lambda - \ep [\log \tau] + \ep[ \tau ] \mathrm{Var}(\mu)) + \lambda \Vert \Lambda - \tilde{\mu} E \Vert_2^2  }{2} \nonumber \\
&& + \frac{ D (\log \lambda - \ep [\log \tau] + \ep[ \tau ] \mathrm{Var}(\mu))  + \lambda \Vert \kappa - \tilde{\mu} \ev \Vert_2^2  }{2}, \nonumber  \vspace{-.35cm}
\end{eqnarray}
where
$\ep[\log \tau] = \psi(\frac{\tilde{\nu}}{2}) + \log \frac{2}{\tilde{S}}$.

\ifCLASSOPTIONcaptionsoff
  \newpage
\fi



%

%
\vspace{-1.2cm}
\begin{IEEEbiography}[{\includegraphics[width=.9in,height=1.1in]{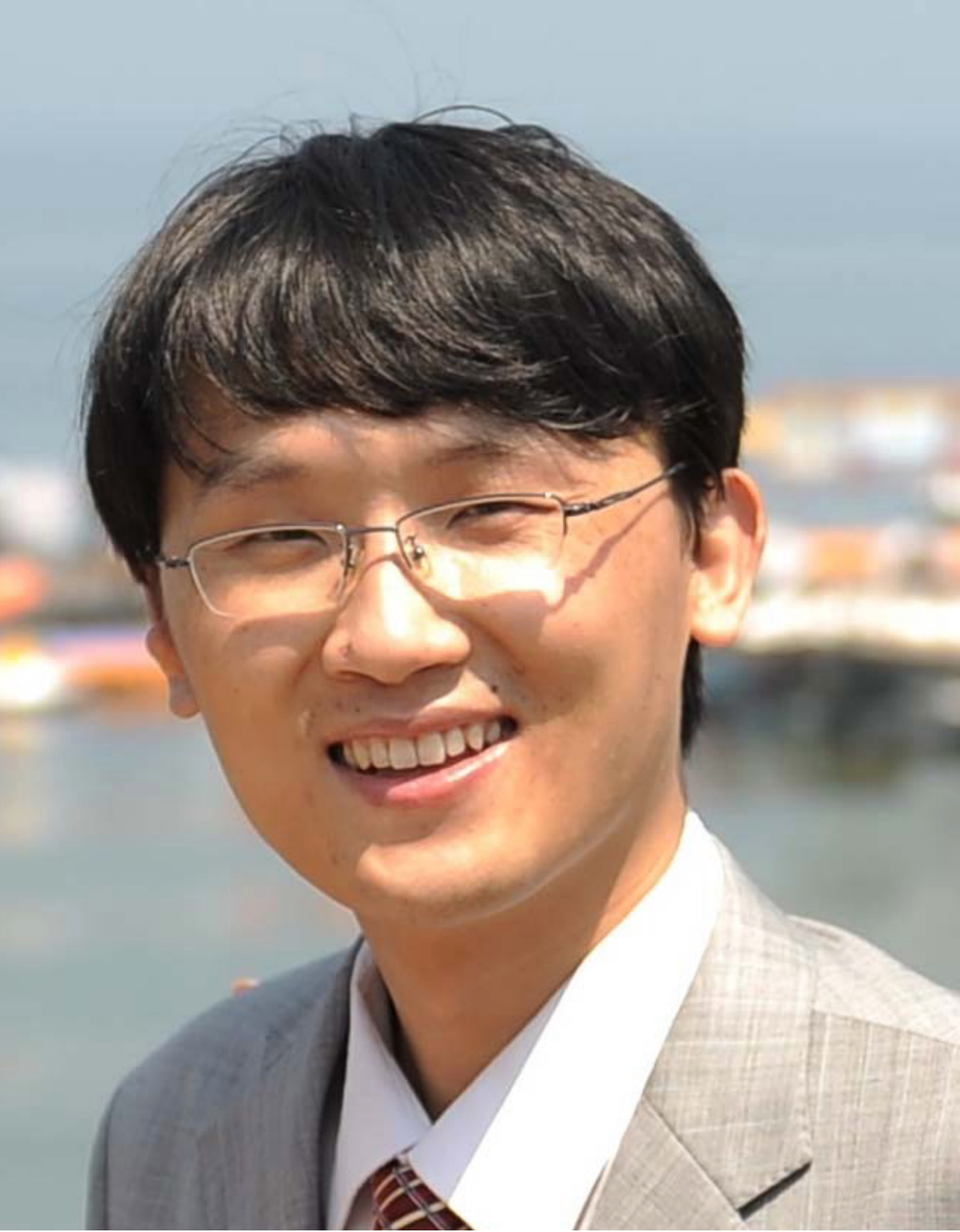}}]{Jun Zhu}
received his BS and PhD degrees from the Department of Computer Science and Technology in Tsinghua University, China, where he is currently an associate professor. He was a project scientist and postdoctoral fellow in the Machine Learning Department, Carnegie Mellon University. His research interests are primarily on machine learning, Bayesian methods, and large-scale algorithms. He was selected as one of the ``AI's 10 to Watch" by IEEE Intelligent Systems in 2013. 
He is a member of the IEEE.
\end{IEEEbiography}\vspace{-1.35cm}


\begin{IEEEbiography}[{\includegraphics[width=.9in,height=1.1in]{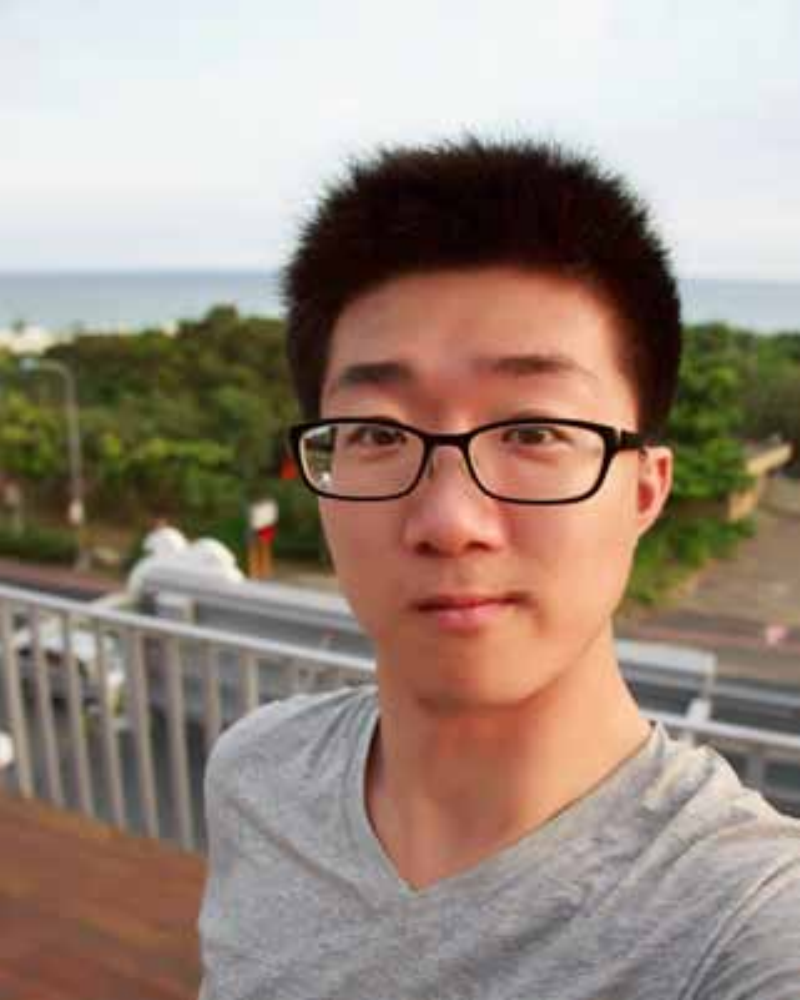}}]{Jiaming Song} is an undergraduate student from Tsinghua University, where he is working as a research assistant in the State Key lab of Intelligent Technology and Systems. His current research interests are primarily on large-scale machine learning, especially Bayesian nonparametrics and deep generative models with applications in social networks and computer vision.
\end{IEEEbiography}\vspace{-1.5cm}

\begin{IEEEbiography}[{\includegraphics[width=.9in,height=1.1in]{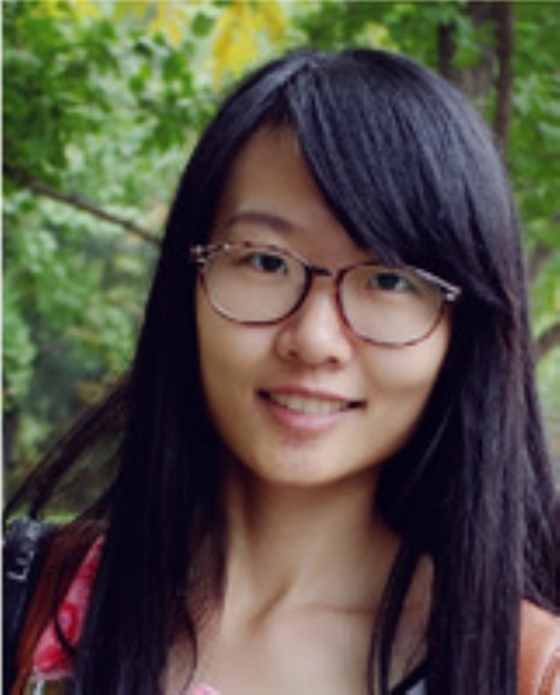}}]{Bei Chen} received her BS from Harbin Institute of Technology, China. She is currently working toward her PhD degree in the Department of Computer Science and Technology at Tsinghua University, China. Her research interests are primarily on machine learning, especially probabilistic graphical models and Bayesian nonparametrics with applications on data mining, such as social networks and discourse analysis.
\end{IEEEbiography}\vspace{-.6cm}




\end{document}